\documentclass[manuscript,screen,nonacm]{acmart} 
\AtBeginDocument{%
  \providecommand\BibTeX{{%
    \normalfont B\kern-0.5em{\scshape i\kern-0.25em b}\kern-0.8em\TeX}}}

\setcopyright{acmcopyright}
\copyrightyear{2021}
\acmYear{2021}
\acmDOI{10.1145/1122445.1122456}

\usepackage{url}
\usepackage{graphicx}
\usepackage{amsmath}
\usepackage{amsthm}
\usepackage{booktabs}
\usepackage{algorithm}
\usepackage{algorithmic}
\usepackage{enumitem}
\usepackage{multicol}
\usepackage{comment}
\usepackage{paralist} 
\usepackage{amsmath,amsfonts} 
\usepackage{breqn}
\usepackage{color, colortbl}
\definecolor{orange}{HTML}{D55E00}
\definecolor{blue}{HTML}{56B4E9}
\definecolor{green}{HTML}{009E73}
\definecolor{purple}{HTML}{882255}

\usepackage{multirow}
\usepackage{nicefrac}
\usepackage{xcolor}
\usepackage{wrapfig}

\usepackage{soul}

\newcommand{\myul}[2][black]{
\setulcolor{#1}\setul{}{1.7pt}\ul{#2}
}\setulcolor{black}

\definecolor{Blue}{rgb}{0, 0, 255}

\usepackage{arydshln}
\usepackage{setspace}

\begin{document}

\title{Generating Fluent Fact Checking Explanations with Unsupervised Post-Editing}

\author{Shailza Jolly}
\affiliation{%
   \institution{TU Kaiserslautern, DFKI GmbH}
   \city{Kaiserslautern}
   \country{Germany}}
\email{shailza.jolly@dfki.de}

\author{Pepa Atanasova}
\affiliation{%
   \institution{Department of Computer Science, University of Copenhagen}
   \city{Copenhagen}
   \country{Denmark}}
\email{pepa@di.ku.dk}

\author{Isabelle Augenstein}
\affiliation{%
   \institution{Department of Computer Science, University of Copenhagen}
   \city{Copenhagen}
   \country{Denmark}}
\email{augenstein@di.ku.dk}


\begin{abstract}
  Fact-checking systems have become important tools to verify fake and misguiding news. These systems become more trustworthy when human-readable explanations accompany the veracity labels. However, manual collection of such explanations is expensive and time-consuming. Recent works frame explanation generation as extractive summarization, and propose to automatically select a sufficient subset of the most important facts from the ruling comments (RCs) of a professional journalist to obtain fact-checking explanations. However, these explanations lack fluency and sentence coherence. In this work, we present an iterative edit-based algorithm that uses only phrase-level edits to perform unsupervised post-editing of disconnected RCs. To regulate our editing algorithm, we use a scoring function with components including fluency and semantic preservation. In addition, we show the applicability of our approach in a completely unsupervised setting. We experiment with two benchmark datasets, LIAR-PLUS and PubHealth. We show that our model generates explanations that are fluent, readable, non-redundant, and cover important information for the fact check.
\end{abstract}






\maketitle

\section{Introduction}
~\label{sec:intro}
In today's era of social media, the spread of news is a click away, regardless of whether it is fake or real. However, the quick propagation of fake news has repercussions on peoples' lives. To alleviate these consequences, independent teams of professional fact checkers manually verify the veracity and credibility of news, which is time and labor-intensive, making the process expensive and less scalable. 
Therefore, the need for accurate, scalable, and explainable automatic fact checking systems is inevitable \cite{kotonya-toni-2020-explainable-automated}.

Current automatic fact checking systems perform veracity prediction for given claims based on evidence documents (\citet{thorne-etal-2018-fever,augenstein-etal-2019-multifc}, \textit{inter alia}), or based on long lists of supporting ruling comments (RCs, \citet{wang-2017-liar,alhindi-etal-2018-evidence}). RCs are in-depth explanations for predicted veracity labels, but they are challenging to read and not useful as explanations for human readers due to their sizable content. 
Recent work \cite{atanasova-etal-2020-generating, kotonya-toni-2020-explainable} has thus proposed to use automatic summarization to select a subset of sentences from long RCs and used them as short layman explanations. However, using a purely extractive approach \cite{atanasova-etal-2020-generating} means sentences are cherry-picked from different parts of the corresponding RCs, and as a result, explanations are often disjoint and non-fluent. 

\begin{figure}[t!]
    \includegraphics[width=1\linewidth]{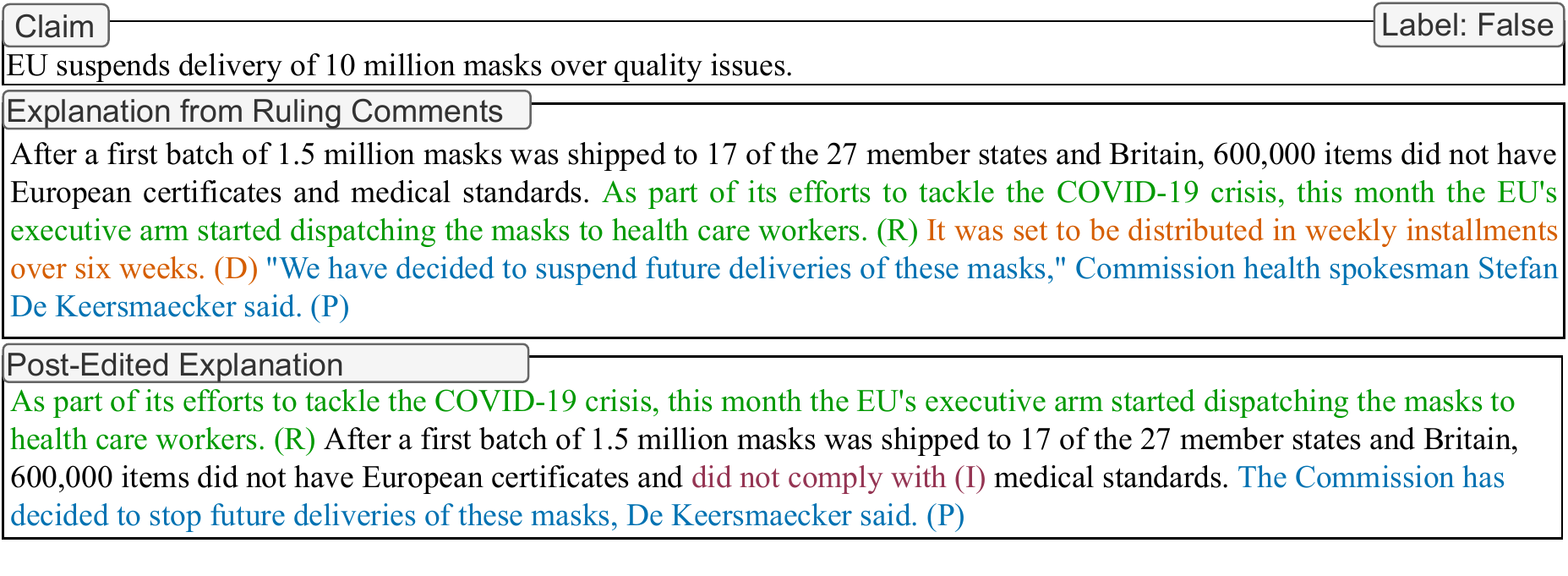}
    \caption{Example of a post-edited explanation from PubHealth that was initially extracted from RCs. We illustrate four post-editing steps: \textcolor{green}{reordering (R)},  \textcolor{purple}{insertion (I)}, \textcolor{orange}{deletion (D)}, and \textcolor{blue}{paraphrasing (P)}.}
    \label{fig:introexp}
\end{figure}

While a sequence-to-sequence model trained on parallel data can partially alleviate these problems, as \citet{kotonya-toni-2020-explainable} propose, it is an expensive affair in terms of the large amount of data and compute required to train these models. Therefore, in this work, we focus on unsupervised post-editing of explanations extracted from RCs. In recent studies, researchers have addressed unsupervised post-editing to generate paraphrases \cite{liu-etal-2020-unsupervised} and sentence simplification \cite{kumar-etal-2020-iterative}. However, they use small single sentences and perform exhaustive word-level or a combination of word and phrase-level edits, which has limited applicability for longer text inputs with multiple sentences, e.g., veracity explanations, due to prohibitive convergence times.

Hence, we present a \textit{novel iterative edit-based algorithm} that performs three edit operations (insertion, deletion, reorder), all at the phrase level. 
Fig.~\ref{fig:introexp} 
illustrates how each post-editing step contributes to creating candidate explanations that are more concise, readable, fluent, and creating a coherent story. 
Our proposed method finds the best post-edited explanation candidate according to a scoring function, ensuring the quality of explanations in fluency and readability, semantic preservation, and conciseness (\S \ref{sec:scoringfunction}). To ensure that the sentences of the candidate explanations are grammatically correct, we also perform grammar checking (\S\ref{sec:gramcorr}). As a second step, we apply paraphrasing to further improve the conciseness and human readability of the explanations (\S\ref{sec:para}). 


In summary, our main \textbf{contributions} are as follows:
\begin{compactitem}
    \item To the best of our knowledge, we are the first to explore an iterative unsupervised edit-based algorithm using only phrase-level edits that leads to feasible solutions for long text inputs.
    \item We show how combining an iterative algorithm with grammatical corrections, and paraphrasing-based post-processing leads to fluent and easy-to-read explanations.
    \item We conduct extensive experiments on the LIAR-PLUS \cite{wang-2017-liar} and PubHealth \cite{kotonya-toni-2020-explainable} fact checking datasets. Our automated evaluation confirms the success of our proposed approach in preserving the semantics important for the fact check and enhancing the readability of the generated explanations. Our manual evaluation confirms that our approach improves the fluency and the conciseness of the generated explanations.
\end{compactitem}

\section{Related Work}
~\label{sec:related}
The most closely related streams of approaches to our work are explainable fact checking, generative approaches to explainability and post-editing for language generation.
\subsection{Explainable Fact Checking}

Recent work has produced fact-checking explanations by highlighting words in tweets using neural attention \cite{lu-li-2020-gcan}. However, their explanations are used only to evaluate and compare the proposed model with other baselines without neural attention. \citet{wu-etal-2020-dtca} propose to model evidence documents with decision trees, which are inherently interpretable ML models. In a recent study, \citet{atanasova-etal-2020-generating} present a multi-task approach to generate free-text explanations for political claims jointly with predicting the veracity of claims. They formulate an extractive summarization task to select a few important sentences from a long fact checking report. \citet{atanasova2021diagnostics} also perform extractive explanation generation guided by a set of diagnostic properties of explanations and evaluate on the FEVER~\cite{thorne-etal-2018-fever} fact checking dataset, where explanation sentences have to be extracted from Wikipedia documents.

In the domain of public health claims, \citet{kotonya-toni-2020-explainable} propose to generate explanations separately from the task of veracity prediction. \citet{mishra-etal-2020-generating} generates summaries of evidence documents from the Web using an attention-based mechanism. They show that their summaries perform better than using the original evidence documents directly. 
Similarly to \citet{atanasova-etal-2020-generating,kotonya-toni-2020-explainable}, we present a generative approach for creating fact checking explanations. In contrast to related work, we propose an unsupervised post-editing approach to improve the fluency and readability of previously extracted fact checking explanations.

\subsection{Generative Approaches to Explainability}
While most work on explanation generation propose methods to highlight portions of inputs (\citet{camburu2018snli,deyoung-etal-2020-eraser}, \textit{inter alia}), some work focuses on generative approaches to explainability. \citet{camburu2018snli} propose combining an explanation generation and a target prediction model in a pipeline or a joint model for Natural Language Inference with abstractive explanations about the entailment of two sentences.
\citet{stammbach2020fever} propose few-shot training for the GPT-3~\cite{brown2020language} model to explain a fact check from retrieved evidence snippets. GPT-3, however, is a limited-access model with high computational costs. As in our work, \citet{kotonya-toni-2020-explainable} first extract evidence sentences, which are then summarised by an abstractive summarisation model. The latter is trained on the PubHealth dataset. In contrast, we are the first to focus on unsupervised post-editing of explanations produced using automatic summarization.


\subsection{Post-Editing for Language Generation}
Previous work has addressed unsupervised post-editing for multiple tasks like paraphrase generation \cite{liu-etal-2020-unsupervised}, sentence simplification \cite{kumar-etal-2020-iterative} or sentence summarization \cite{schumann-etal-2020-discrete}. However, all these tasks handle shorter inputs in comparison to the long multi-sentence extractive explanations that we have. Furthermore, they perform exhaustive edit operations at the word level and sometimes additionally at the phrase level, both of which increase computation and inference complexity.
Therefore, we present a novel approach that performs a fixed number of edits only at the phrase level followed by grammar correction and paraphrasing.

\section{Method}
~\label{sec:method}
Our method is comprised of two steps. First, we select sentences from RCs that serve as extractive explanations for verifying claims (\S\ref{sec:method:selection}). We then apply a post-editing algorithm on the extractive explanations in order to improve their fluency and coherence (\S\ref{sec:method:post-editing}). 

\subsection{Selecting Sentences for Post-Editing}
~\label{sec:method:selection}
\textbf{Supervised Selection.} 
To produce supervised extractive explanations, we build models based on DistilBERT~\cite{Sanh2019DistilBERTAD} for LIAR-PLUS, and SciBERT~\cite{beltagy-etal-2019-scibert} for PubHealth to allow for direct comparison with \citet{atanasova-etal-2020-generating,kotonya-toni-2020-explainable}. 
We supervise explanation generation by $k$ greedily selected sentences from each claim's RCs that achieve the highest ROUGE-2 F1 score when compared to the gold justification. We choose $k=4$ for LIAR-PLUS and $k=3$ for PubHealth, the average number of sentences in the veracity justifications in the corresponding datasets. The selected sentences are positive gold labels, $\mathbf{y}^E \in \{0,1\}^N $, where $N$ is the number of sentences in the RCs. We also use the veracity labels $\mathbf{y}^{F} \in Y_{F}$ for supervision. Following~\citet{atanasova-etal-2020-generating-fact}, we then learn a multi-task model $g(X) = (\mathbf{p}^E, \mathbf{p}^F)$. Given the input X, comprised of a claim and the RCs, it predicts jointly the veracity explanation $\mathbf{p}^E$ and the veracity label $\mathbf{p}^F$, where $\mathbf{p}^E \in \mathbb{R}^{1, N}$ selects sentences for explanation, i.e. \{0,1\}, and  $\mathbf{p}^{F} \in \mathbb{R}^{m}$, with $m=6$ for LIAR-PLUS, and $m=4$ for PubHealth. Finally, we optimise the joint cross-entropy loss function $\mathcal{L}_{MT}= \mathcal{H}(\mathbf{p}^{E}, \mathbf{y}^{E}) + \mathcal{H}(\mathbf{p}^{F}, \mathbf{y}^{F})$.

\textbf{Unsupervised Selection.} 
We also experiment with unsupervised selection of sentences to test the possibility to construct fluent fact checking explanations in an entirely unsupervised way. We use a Longformer~\cite{Beltagy2020Longformer} model, which was introduced for tasks with longer input, instead of the sliding-window approach also used in \citet{atanasova-etal-2020-generating-fact}, which is without cross-window attention. 
We train a model $h(X)\!=\!\mathbf{p}^F$ to predict the veracity of a claim. We optimise a cross-entropy loss function $\mathcal{L}_{F}\!=\!\mathcal{H}(\mathbf{p}^{F}, \mathbf{y}^{F})$ and select $k$ sentences $\mathbf{p}^{E'}\!\in\!\mathbb{R}^{1, N}$, \{0, 1\}, with the highest saliency scores. The saliency score of a sentence is the sum of the saliency scores of its tokens. The saliency of a token is the gradient of the input token w.r.t. the output~\cite{Simonyan2014DeepIC}. We selected sentences using the raw gradients as \citet{atanasova-etal-2020-diagnostic} show that different gradient-based methods yield similar results. As the selection could be noisy~\cite{kindermans2019reliability}, we consider these experiments as only complementary to the main supervised results.

\subsection{Post-Editing}~\label{sec:method:post-editing}
Our post-editing is completely unsupervised and operates on sentences obtained in Sec.~\ref{sec:method:selection}. It is a search algorithm that evaluates the candidate sequence $\mathbf{p}^{C}$ for a given input sequence, where the input sequence is either $\mathbf{p}^{E}$ for supervised selection or $\mathbf{p}^{E'}$ for unsupervised selection. Below, we use $\mathbf{p}^{E}$ as a representative of both $\mathbf{p}^{E}$ and $\mathbf{p}^{E'}$.

Given $\mathbf{p}^{E}$, we iteratively generate multiple candidates by performing phrase-level edits (\S\ref{sec:cadidategeneration}). To evaluate a candidate explanation, we define a scoring function, which is a product of multiple scorers, also known as a product-of-experts model \cite{hinton2002training}. Our scoring function includes fluency and semantic preservation, and controls the length of the candidate explanation (\S\ref{sec:scoringfunction}). 
We repeat the process for $n$ steps and select the last best-scoring candidate as our final output. We then use grammar correction (\S\ref{sec:gramcorr}) and paraphrasing (\S\ref{sec:para}) to further ensure conciseness and human readability.

\subsubsection{Candidate sequence generation}
~\label{sec:cadidategeneration}
We generate candidate sequences by phrase-level edits. 
We use the off-the-shelf syntactic parser from CoreNLP \cite{manning-etal-2014-stanford} to obtain the constituency tree of a candidate sequence $\mathbf{p}^{C}$. As $\mathbf{p}^{C}$ is long, we perform all operations at the phrase level. 
At each step $t$, our algorithm first randomly picks one operation -- insertion, deletion, or reordering, and then randomly selects a phrase. 

For \textbf{insertion}, our algorithm inserts a <MASK> token before the randomly selected phrase, and uses RoBERTa to evaluate the posterior probability of a candidate word \cite{NEURIPS2020_7a677bb4}. This allows us to leverage the pre-training capabilities of RoBERTa and inserts high-quality words that support the context of the overall explanation. 
Furthermore, inserting a <MASK> token before a phrase prevents breaking other phrases within the explanation, thus preserving their fluency. 

The \textbf{deletion} operation deletes the randomly selected phrase.
For the \textbf{reorder} operation 
we randomly select one phrase, 
which we call \textit{reorder phrase}, and randomly select $m$ phrases, which we call \textit{anchor phrases}. 

We \textbf{reorder} each \textit{anchor phrase} with a \textit{reorder phrase} and obtain $m$ candidate sequences. We feed these candidates to GPT2 and select the most fluent candidate based on the fluency score given by Eq. \ref{eq:fluency}.

\subsubsection{Scoring Functions}
~\label{sec:scoringfunction}
The \textbf{fluency score} ($f_{flu}$) measures the language fluency of a candidate sequence. 
We use pre-trained GPT2 model \cite{radford2019language}. We use the joint likelihood of candidate $\mathbf{p}^{C}$:
\begin{equation}
f_{flu}(\mathbf{p}^{C}) = \prod\nolimits_{i = 1}^{n} P(\mathbf{p}^{C}_i|\mathbf{p}^{C}_1, ...., \mathbf{p}^{C}_{i-1})
\label{eq:fluency}
\end{equation}

In Sec.~\ref{sec:results:manual:quality} we evaluate the achieved fluency of the generated explanations through human evaluation. Additionally, as the fluency score measures the likelihood of the text according to GPT2, which is trained on 40GB of Internet text, we assume that complex text that is not common or is not likely to appear on the Internet, would also have lower fluency score. Hence, we expect that improving the fluency of an explanation, would lead to explanations that are more easily understood. We evaluate the latter in Sec.~\ref{sec:result:readability} through the automated readability scores.

\textbf{Length score ($f_{len}$)} This score encourages the generation of shorter sentences. We assume that reducing the length of the generated explanation is also beneficial for improving the readability of the explanation as it promotes shorter sentences, which are easier to read. It is proportional to the inverse of the sequence length, i.e., the higher the length of a candidate sentence, the lower its score. To control over-shortening, we reject explanations with fewer than 40 tokens. The number of tokens is a hyperparameter that we chose after fine-tuning on the validation split.

For \textbf{semantic preservation}, we compute similarities at both word and explanation level between our source explanation ($\mathbf{p}^{E}$) and candidate sequence ($\mathbf{p}^{C}$) at time-step $t$. The word-level semantic scorer evaluates the preserved amount of keyword information in the candidate sequence. Similarly to \citet{NEURIPS2020_7a677bb4}, we use RoBERTa (R) \cite{liu2019roberta}, a pre-trained masked language model, to compute a contextual representation of the ith word in an explanation as R$(\mathbf{p}^{E}_i, \mathbf{p}^{E})$. Here, $\mathbf{p}^{E}\!=\!(\mathbf{p}^{E}_1\dots \mathbf{p}^{E}_m)$ is an input sequence of words. We then extract keywords from $\mathbf{p}^{E}$ using Rake \cite{rose2010automatic} and compute a \textbf{keyword-level semantic similarity score}:
\begin{equation}
\small
f_{w}(\mathbf{p}^{E}, \mathbf{p}^{C})\!=\!\min_{k \in kw(\mathbf{p}^{E})}
\max_{\mathbf{p}^{C}_i \in \mathbf{p}^{C}} R(k, \mathbf{p}^{E})^\intercal R(\mathbf{p}^{C}_i, \mathbf{p}^{C})
\label{eq:wordsemantics}
\end{equation}
which is the lowest cosine similarity among all keywords i.e. the least matched keyword of $\mathbf{p}^{E}$.

The keyword-level semantic similarity accounts for preserving the semantic information of the separate keywords used in the text. It is, thus, not affected by changes in words that do not bear significant meaning for the overall explanation. However, as this semantic similarity is performed at keyword-level it does not account for preserving the overall meaning of the text and the context that the keywords are used in. 

Hence, we also employ a \textbf{explanation-level semantic preservation scorer} that measures the cosine similarity of two explanation vectors, where the explanation vectors are explanation encodings that contain the overall semantic meaning of the explanation: 
\begin{equation}
f_{e}(\mathbf{p}^{E}, \mathbf{p}^{C}) = \tfrac{(\mathbf{p}^{C})^\intercal  \mathbf{p}^{E}}{||\mathbf{p}^{C}||\mathbf{p}^{E}||}
\label{eq:sentsemantics}
\end{equation}
\noindent We use SBERT \cite{reimers-gurevych-2019-sentence} for obtaining embeddings for both $\mathbf{p}^{E}$, $\mathbf{p}^{C}$. Our overall semantic score is the product of the word level and the explanation level semantics scores:
\begin{equation}
\small
f_{sem}(\mathbf{p}^{E}, \mathbf{p}^{C}) = f_{w}(\mathbf{p}^{E}, \mathbf{p}^{C})^\beta . \\ f_{e}(\mathbf{p}^{E}, \mathbf{p}^{C})^\eta
\label{eq:overallsemantics}
\end{equation}
\noindent where $\beta$, and $\eta$ are hyperparameter weights for the separate scores. We evaluate the semantic preservation of the post-edited explanations with automated ROUGE scores (\S\ref{sec:result:rouge}) and manual human annotations (\S\ref{sec:results:manual:quality}, \S\ref{sec:results:manual:info}).

Lastly, the \textbf{Named Entity (NE) score ($f_{ent}$)} is an additional approximation we include as a measure of meaning preservation, since NEs hold the key information within a sentence. We first identify NEs using an off-the-shelf entity tagger\footnote{https://spacy.io/} and then count their number in a given explanation. 

Our \textbf{overall scoring} function is the product of individual scores, where $\alpha$, $\gamma$, and $\delta$ are hyperparameter weights for the different scores:
\begin{equation}
\small
f_(\mathbf{p}^{C}) = f_{flu}(\mathbf{p}^{C})^\alpha .  f_{sem}(\mathbf{p}^{E}, \mathbf{p}^{C}). \\ f_{len}(\mathbf{p}^{C})^\gamma .  f_{ent}(\mathbf{p}^{C})^\delta 
\label{eq:overallscore}
\end{equation}

\subsubsection{Iterative Edit-based Algorithm}
~\label{sec:iterativeedit}
Given input explanations, our algorithm iteratively performs edit operations for $n$ steps to search for a highly scored candidate ($\mathbf{p}^{C}$). At each search step, it computes scores for the previous sequence ($\mathbf{p}^{C-1}$) and candidate sequence using Eq. \ref{eq:overallscore}. It selects $\mathbf{p}^{C}$ if its score is larger than $\mathbf{p}^{C-1}$ by a multiplicative factor $r_{op}$:
\begin{equation}
\nicefrac{f_{\mathbf{p}^{C}}}{f_{\mathbf{p}^{C-1}}} > r_{op}
\label{eq:selectop}
\end{equation}

\noindent For each edit operation, we use a separate threshold value $r_{op}$. 
$r_{op}$ allows controlling specific operations where $r_{op} < 1$ allows the selection of candidates ($\mathbf{p}^{C}$) which have lower scores than $\mathbf{p}^{C-1}$.
We tune all hyperparameters, including $r_{op}$, $n$, etc., using the validation split of the LIAR-PLUS dataset.

\subsubsection{Grammatical Correction}
~\label{sec:gramcorr}
Once the best candidate explanation is selected, we feed it to a language toolkit\footnote{https://github.com/jxmorris12/language_tool_python}, which detects grammatical errors like capitalization and irrelevant punctuation, and returns a corrected version of the explanation. Furthermore, to ensure that we have no incomplete sentences, we remove sentences without verbs in the explanation. These two steps further ensure that the generated explanations are fluent (further evaluated in Sec.~\ref{sec:results:manual:quality}). 

\subsubsection{Paraphrasing}
~\label{sec:para}
Finally, to improve fluency and readability further, we use Pegasus \cite{zhang2020pegasus}, a model pre-trained with an abstractive text summarization objective. It focuses on relevant input parts to summarise the input semantics in a concise and readable way. Since we want our explanations to be both fluent and human-readable, we leverage Pegasus without fine-tuning on downstream tasks. 
This way, after applying our iterative edit-based algorithm with grammatical error correction and paraphrasing, we obtain explanations that are fluent, coherent, and human readable. 

\section{Experiments}
\subsection{Datasets}

\begin{table}[t]
\centering
\begin{tabular}{lrrr}
\toprule
\textbf{Dataset} & \textbf{Train size} & \textbf{Dev size} & \textbf{Test size} \\ \midrule
LIAR-PLUS & 10,146 & 1,278 & 1,255 \\
PubHealth & 9,817 & 1,227 & 1,235 \\
\bottomrule
\end{tabular}
\caption{Size of the fact checking datasets used in this work (\S\ref{sec:dataset}).}
\label{tab:datasets}
\end{table}

\label{sec:dataset}
We use two fact checking datasets, LIAR-PLUS~\cite{wang-2017-liar} and PubHealth~\cite{kotonya-toni-2020-explainable}. These are the only two available real-world fact checking datasets that provide short veracity justifications along with claims, RCs, and veracity labels. Table~\ref{tab:datasets} provides the size for each of the splits in the corresponding dataset. The labels used in LIAR-PLUS are \{true, false, half-true, barely-true, mostly-true, pants-on-fire\}, and in PubHealth, \{true, false, mixture, unproven\}.
While claims in LIAR-PLUS are only from PolitiFact, PubHealth contains claims from eight fact checking sources. PubHealth has also been manually curated, e.g., to exclude poorly defined claims. Finally, the claims in PubHealth are more challenging to read than those in LIAR-PLUS and other real-world fact checking datasets.

\subsection{Models}
Our experiments include the following models; their hyperparameters are given in Appendix \ref{sec:experiments}.

\textbf{(Un)Supervised Top-N} extracts sentences from the RCs, which are later used as input to our algorithm. The sentences are extracted in either a supervised or unsupervised way (\S\ref{sec:method:selection}).

\textbf{(Un)Supervised Top-N+Edits-N} generates explanations with the iterative edit-based algorithm (\S\ref{sec:iterativeedit}) and grammar correction (\S\ref{sec:gramcorr}). The model is fed with sentences extracted from RCs in an (un)supervised way.

\textbf{(Un)Supervised Top-N+Edits-N+Para} generates explanations by paraphrasing the explanations produced by (Un) Supervised Top-N+Edits-N (\S\ref{sec:para}).


\textbf{\citet{atanasova-etal-2020-generating-fact}} is a reference model that trains a multi-task system to predict veracity labels and extract explanation N sentences, where N is the average number of the sentences in the justifications of each dataset.
\citet{kotonya-toni-2020-explainable} is a baseline model that generates abstractive explanations with an average sentence length of 3.

\textbf{Lead-K}~\cite{nallapati2017summarunner} is a common lower-bound baseline for summarisation models. It selects the first K sentences of the RCs.
\subsection{Evaluation Overview}
We perform both automatic and manual evaluations of the models above. We include automatic measures for assessing readability (\S\ref{sec:result:readability}). While the latter was not included in prior work, we consider readability an essential quality of an explanation, and thus report it. We further include automatic ROUGE F1 scores (overlap of the generated explanations with the gold ones, \S\ref{sec:result:rouge}) for compatibility with prior work and to ensure that our generated explanations don't shift much from the gold ones. In particular, we are interested whether the reported ROUGE scores for the post-edited explanations are not significantly different from the ROUGE scores of the non-edited explanations, which would indicate a preservation of the original content important for the fact check. We note, however, that the employed automatic measures are limited as they are based on word-level statistics. Especially ROUGE F1 scores should be taken with a grain of salt, as only exact matches of words are rewarded with higher scores, where paraphrases or synonyms of words in the gold summary are not scored. 
Hence, we also conduct a manual evaluation following~\citet{atanasova-etal-2020-generating-fact} to further assess the quality of the generated explanations with a user study. As manual evaluation is expensive to obtain, the latter is, however, usually estimated based on small samples.

\section{Automatic Evaluation}
~\label{sec:result:automated}
As mentioned above, we use ROUGE F1 scores to compute overlap between the generated explanations and the gold ones, and compute readability scores to assess how challenging the produced explanations are to read.

\subsection{Readability Results}
~\label{sec:result:readability}

\textbf{Metrics.} Readability is a desirable property for fact checking explanations, as explanations that are challenging to read would fail to convey the reasons for the chosen veracity label and would not improve the trust of end-users. To evaluate 
readability, we compute Flesch Reading Ease~\cite{kincaid1975derivation} and Dale-Chall Readability Score~\cite{powers1958recalculation}. The Flesch Reading Ease metric gives a text a score $\in [1, 100]$, where a score $\in [30, 50]$ requires college education and is difficult to read, a score $\in (50, 60]$ requires a 10th to 12th school grade and is still fairly difficult to read, a score $\in (60, 70]$ is regarded as plain English, which is easily understood by 13- to 15-year-old students. The Dale-Chall Readability Score uses a specially designed list of words familiar to lower-grade students to assess the number of hard words used in a given text. It gives a text a score $\in [9.0, 9.9]$ when it is easily understood by a 13th to 15th-grade (college) student, a score $\in [8.0, 8.9]$ when it is easily understood by an 11th or 12th-grade student, a score $\in [7.0, 7.9]$ when it is easily understood by a 9th or 10th-grade student. The scores presented are an average over the readability scores for the separate instances in the test split (see supplemental material for results on the validation split). We additionally provide the 95\% confidence interval for the average score based on 1000 random re-samples from the corresponding split.

\textbf{Results.} Table~\ref{tab:results:readability:main} presents the readability results. We find that our iterative edit-based algorithm consistently improves the reading ease of the explanations by up to 5.16 points, and reduces the grade requirement by up to 0.30 points. Furthermore, the improvements are statistically significant ($p$<0.05) in both supervised and unsupervised explanations, except for the Dale-Chall score for the LIAR unsupervised explanations, where the 95\% confidence interval is still decreased compared to the non-edited explanations. Conducting paraphrasing further improves significantly ($p$<0.05) the reading ease of the text by up to 9.33 points, and reduces the grade requirement by up to 0.48 points. It is also worth noting that the explanations produced by \citet{atanasova-etal-2020-generating-fact} as well as the gold justifications are fairly difficult to read and can require even college education for grasping the explanation, while the explanations generated by our algorithm can be easily understood by 13- to 15-year-old students according to the Flesch Reading Ease score.

\textbf{Overall observations.} Our results show that our method makes fact checking explanations less challenging to read and makes them accessible to a broader audience of up to 10th-grade students.
\bgroup
\def\arraystretch{1.2}

\begin{table*}[t]
\centering
\fontsize{10}{10}\selectfont
\begin{tabular}{llrrrr}
\toprule
& \bf Method &  {\bf Flesch $\nearrow$} &  {\bf Flesch  CI} & {\bf Dale-Chall $\searrow$}  & {\bf Dale-Chall CI}  \\
\midrule
\multicolumn{6}{c}{\bf LIAR-PLUS}\\ 
\multirow{2}{*}{\bf Baselines} & Lead-4 & 51.70 & { [50.93-52.53]} & 8.73 & { [8.67-8.78]} \\
& Lead-6 &  53.24 &  { [52.58-53.92]} & 8.43  & { [8.38-8.47]} \\
\hdashline
\multirow{3}{*}{\bf Supervised} & Top-6 (Supervised) & 58.82  & { [58.13-59.54]} & 7.88 &  { [8.17-8.28]} \\
& Top-6+Edits-6 & 60.21 &  { [59.51-60.95]} & 7.75  &   { [7.70-7.80]} \\
& Top-6+Edits-6+Para & 66.34 &  { [65.73-66.98]} & 7.42  & { [7.37-7.47]} \\
\hdashline
\multirow{3}{*}{\bf Unsupervised} & Top-6 (Unsupervised) & 53.33  & { [52.70-53.92]} & 8.50 &  { [8.46-8.54]}  \\
& Top-6+Edits-6 & 55.25 &  { [54.60-55.88]} & \textcolor{purple}{8.46} &  { [8.42-8.51]} \\
& Top-6+Edits-6+Para & 62.13 &  { [61.56-62.71]} & 8.11 &  { [8.06-8.15]} \\
\hdashline
 & \citet{atanasova-etal-2020-generating}-4 & 58.56 &  { [57.75-59.31]} & 7.99  &  {[7.94-8.03]}\\
& Justification & 58.81 &  { [58.22-59.41]} & 8.23 &  { [7.93-8.04]} \\
\midrule
\multicolumn{6}{c}{\bf PubHealth} \\
& Lead-3 & 44.44 &  { [43.05-45.68]} & 9.12  & { [9.05-9.19]} \\
& Lead-5 & 45.96 &  { [44.80-46.98]} & 8.85  & { [8.79-8.91]} \\
\hdashline
\multirow{3}{*}{\bf Supervised} & Top-5 (Supervised) & 48.63  & { [47.91-49.44]} & 8.67 &  { [8.62-8.72]}  \\
& Top-5+Edits-5 & 53.79 &  { [53.01-54.56]} & 8.37  & { [8.31-8.42]} \\
& Top-5+Edits-5+Para & 61.39 &  { [60.71-62.10]} & 7.97  & { [7.92-8.03]} \\
\hdashline
\multirow{3}{*}{\bf Unsupervised} & Top-5 (Unsupervised) & 45.20  & { [44.41-46.04]} & 8.94 &  { [8.89-8.98]} \\
& Top-5+Edits-5 & 50.74 &  { [49.89-51.53]} & 8.63 &  { [8.57-8.68]} \\
& Top-5+Edits-5+Para & 60.07 &  { [59.37-60.77]} & 8.15 &  { [8.09-8.20]} \\
\hdashline
& \citet{atanasova-etal-2020-generating}-3 & 48.73 &  { [47.81-49.66]} & 8.88  & { [8.82-8.94]} \\
& Justification & 49.29 &  { [48.27-50.40]} & 9.17  & { [9.08-9.26]} \\
\bottomrule
\end{tabular}
\caption{Readability measures (\S\ref{sec:result:readability}) over the test splits (for validation and ablations, see the appendix). Readability measures include 95\% confidence intervals (\S\ref{sec:result:readability}, Metrics.). We report results reported from the prior work of \citet{atanasova-etal-2020-generating}-N, where we have the outputs to compute readability. Readability scores for Top-N+Edits-N and  Top-N+Edits-N+Para are statistically significant ($p$<0.05) compared to Top-N, and to \citet{atanasova-etal-2020-generating}-3/4, except for the score in \textcolor{purple}{purple}.}
\label{tab:results:readability:main}
\end{table*}

\subsection{Automatic ROUGE Scores}
~\label{sec:result:rouge}

\begin{table*}[t]
\centering
\fontsize{10}{10}\selectfont
\begin{tabular}{llrrrrrrrr}
\toprule
& \bf Method & \textbf{R-1}$\nearrow$ & \textbf{R-1 CI}  & \textbf{R-2}$\nearrow$ & \textbf{R-2 CI} &  \textbf{R-L}$\nearrow$ & \textbf{R-L CI}  \\
\midrule
\multicolumn{9}{c}{\bf LIAR-PLUS}\\ 
\multirow{2}{*}{\bf Baselines} & Lead-4 & \textit{28.11} & {\small [27.39-28.39]}& \textit{6.96} &{\small  [6.52\hspace{1.25mm}-\hspace{1.25mm}7.33]}& \textit{24.38} & {\small [23.73-24.68]}  \\
& Lead-6 &  29.15 & {\small [28.66-29.69]} & 8.28 & {\small [7.85\hspace{1.25mm}-\hspace{1.25mm}8.67]} & 25.84 & {\small [25.35-26.30]} \\
\hdashline
\multirow{3}{*}{\bf Sup.} & Top-6 (Supervised) & 34.42 & {\small [33.78-35.00]}&12.36 & {\small [11.85-12.84]} & 30.58 & {\small [30.01-31.13]}\\
& Top-6+Edits-6 & 33.92 & {\small [33.31-34.53]} & 11.73 & {\small [11.29-12.24]} & 30.01 & {\small [29.43-30.60]} \\
& Top-6+Edits-6+Para & 33.94 &{\small  [33.37-34.47] }& \underline{11.25} & {\small [10.81-11.73] }& 30.08 & {\small [29.49-30.59] }\\
\hdashline
\multirow{3}{*}{\bf Unsup.} & Top-6 (Unsupervised) & 29.63 & {\small [29.03-30.07] }& 7.58 &{\small  [7.53\hspace{1.25mm}-\hspace{1.25mm}8.25]} & 25.86 &{\small  [25.52-26.47]} \\
& Top-6+Edits-6 & 28.93 & {\small [28.42-29.42]} & \underline{7.06} & {\small [6.74\hspace{1.25mm}-\hspace{1.25mm}7.43]} & 25.14 & {\small [24.69-25.61]} \\
& Top-6+Edits-6+Para & 28.98 & {\small [28.50-29.52]}& \underline{6.84} & {\small [6.51\hspace{1.25mm}-\hspace{1.25mm}7.16]} &  25.39 & {\small [24.95-25.87]}\\
\hdashline
 & \citet{atanasova-etal-2020-generating}-4 & \textit{35.70} & {\small [34.23-35.39]} & \textit{13.51} & {\small [12.47-13.67]} & \textit{31.58} & {\small [30.07-31.21]} \\
\midrule

\multicolumn{9}{c}{\bf PubHealth} \\
\multirow{3}{*}{\bf Baselines} & Lead-3 & \textit{29.01} & - & \textit{10.24} & - & \textit{24.18} & - \\
& Lead-3 & 23.05 & {\small [22.53-23.59]} & 6.28 & {\small [5.48\hspace{1.25mm}-\hspace{1.25mm}6.37]} & 19.27 & {\small [18.42-19.40]} \\
& Lead-5 & 23.73 & {\small [22.50-23.62]} & 6.86 & {\small [5.81\hspace{1.25mm}-\hspace{1.25mm}6.58]} & 20.67 & {\small [19.08-20.18]}\\
\hdashline
\multirow{3}{*}{\bf Sup.} & Top-5 (Supervised) & 29.93 & {\small [28.87-30.97]} & 12.42 & {\small [11.44-13.63]} & 26.24 & {\small [25.21-27.44]}  \\
& Top-5+Edits-5 & 29.38 & {\small [28.45-30.33]} & 11.16 & {\small [10.17-12.15]} & 25.41 & {\small [24.48-26.41]} \\
& Top-5+Edits-5+Para & 28.40 & {\small [27.55-29.17]} & \underline{9.56} & {\small [8.89\hspace{0.7mm}-\hspace{0.7mm}10.23]} & \underline{24.37} &{\small  [23.52-25.10]} \\
\hdashline
\multirow{3}{*}{\bf Unsup.} & Top-5 (Unsupervised) & 23.52 & {\small [22.95-24.12]} & 6.12 & {\small [5.76\hspace{1.25mm}-\hspace{1.25mm}6.46]}& 19.93 &{\small  [19.37-20.45]} \\
& Top-5+Edits-5 & 23.09 & {\small [22.55-23.64]} & 5.56 & {\small [5.24\hspace{1.25mm}-\hspace{1.25mm}5.92]} & 19.44 &{\small  [18.93-19.93]}  \\
& Top-5+Edits-5+Para & 23.35 & {\small [22.85-23.86] }& \underline{5.38} & {\small [5.09\hspace{1.25mm}-\hspace{1.25mm}5.71]} & 19.56 & {\small [19.08-20.03]} \\
\hdashline
& \citet{kotonya-toni-2020-explainable}-3 & \textit{32.30} & - & \textit{13.46} & - & \textit{26.99} & -\\
& \citet{atanasova-etal-2020-generating}-3 & 33.55 & {\small [29.79-31.65]} & 13.12 &{\small  [11.17-13.42]} & 29.41 &{\small  [25.27-27.31]}\\
\bottomrule
\end{tabular}
\caption{ROUGE-1/2/L F1 scores (\S\ref{sec:result:rouge}) of supervised (Sup.) and usupervised (Unsup.) methods over the test splits (for validation and ablations, see the appendix). In \textit{italics}, we report results reported from prior work, where we do not always have the outputs to compute the confidence intervals. \underline{Underlined} ROUGE scores of the Top-N+Edits-N and Top-N+Edits-N+Para are statistically significant ($p<0.05$) compared to the input Top-N ROUGE scores, N=\{5,6\}.}
\label{tab:results:rouge:main}
\end{table*}

\bgroup
\def\arraystretch{1}

\textbf{Metrics.} To evaluate the generated explanations w.r.t. the gold justifications, we follow~\citet{atanasova-etal-2020-generating-fact, kotonya-toni-2020-explainable} and use measures from automatic text summarisation -- ROUGE-1, ROUGE-2, and ROUGE-L F1 scores. These account for n-gram (1/2) and longest (L) overlap between generated and gold justification. The scores are recall-oriented, i.e., they calculate how many of the n-grams in the gold text appear in the generated one.

\textbf{Caveats.} Here, ROUGE scores are used to verify that the generated explanations preserve information important for the fact check, as opposed to generating completely unrelated text. Thus, we are interested in whether the ROUGE scores of the post-edited explanations are close but not necessarily higher than those of the selected sentences from the input RCs. It is worth noting, we include paraphrasing and insertion of new words to improve explanation's readability, which, while bearing the same meaning, necessarily results in lower ROUGE scores.

\textbf{Results.} In Table~\ref{tab:results:rouge:main}, we present the ROUGE score results. First, comparing the results for the input Top-N sentences with the intermediate and final explanations generated by our system, we see that, while very close, the ROUGE scores tend to decrease. For PubHealth, we also see that the intermediate explanations always have higher ROUGE scores compared to the final explanations from our system. These observations corroborate two main assumptions about our system. First, our system manages to preserve a large portion of the information important for explaining the veracity label, which is also present in the justification. This is further corroborated by observing that the decrease in the ROUGE scores is often not statistically significant ($p<0.05$, except for some ROUGE-2 and one ROUGE-L score). Second, the operations in the iterative editing and the subsequent paraphrasing allow for the introduction of novel n-grams, which, while preserving the meaning of the text, are not explicitly present in the gold justification, thus, affecting the word-level ROUGE scores. We further discuss this in Sec.~\ref{sec:discussion} and the appendix.

The ROUGE scores of the explanations generated by our post-editing algorithm when fed with sentences selected in an unsupervised way are considerably lower than with the supervised models. The latter illustrates that supervision for extracting the most important sentences is important to obtain explanations close to the gold ones. Finally, the systems' results are mostly above the LEAD-N scores, with a few exceptions for the unsupervised explanations for LIAR-PLUS.   

\textbf{Overall observations.} We note that while automatic measures can serve as sanity checks and point to major discrepancies between generated explanations and gold ones, related work in generating fact checking explanations~\cite{atanasova-etal-2020-generating-fact} has shown that the automatic scores to some extent disagree with human evaluation studies, as they only capture word-level overlap and cannot reflect improvements of explanation quality. Human evaluations are therefore conducted for most summarisation models~\cite{chen-bansal-2018-fast,tan-etal-2017-abstractive}, which we include in Sec.~\ref{sec:results:manual}.

\section{Manual Evaluation}~\label{sec:results:manual}
As automated ROUGE scores only account for word-level similarity between the generated and the gold explanation, and the readability scores account only for surface-level characteristics of the explanation, we further conduct a manual evaluation of the quality of the produced explanations.

\subsection{Explanation Quality}~\label{sec:results:manual:quality}
We manually evaluate two explanations: the input Top-N sentences,  and the final explanations produced after paraphrasing (Edits-N+Para). We perform a manual evaluation of the test explanations obtained from supervised selection for both datasets with two annotators for each. Both annotators have a university-level education in English. 

\textbf{Metrics.} We show a claim, veracity label, and two explanations to each annotator and ask them to rank the explanations according to the following criteria. \textbf{Coverage} means the explanation contains important and salient information for the fact check. \textbf{Non-redundancy} implies the explanation does not contain any redundant/repeated/not relevant information to the claim and the fact check. \textbf{Non-contradiction} checks if there is information contradictory to the fact check. \textbf{Fluency} measures the grammatical correctness of the explanation and if there is a coherent story. \textbf{Overall} measures the overall explanation quality. We allow annotators to give the same rank to both explanations~\cite{atanasova-etal-2020-generating-fact}. We randomly sample 40 instances\footnote{We select 40 instances due to the complexity of the annotation task, which increases its cost and execution time, and also following related work that manually evaluates fact checking explanations~\citet{atanasova-etal-2020-generating-fact} and related work that manually evaluates machine-generated summaries~\cite{liu-lapata-2019-text}.} and do not provide the annotators with information about the explanation type.

\textbf{Results.} Table \ref{tab:results:humanevals_task1} presents the human evaluation results for the first task. Each row indicates the annotator number and the number of times they ranked an explanation higher for one criterion. 
Our system's explanations achieve higher acceptance for non-redundancy and fluency for LIAR-PLUS. The results are more pronounced for the PubHealth dataset, where our system's explanations were preferred in almost all metrics by both annotators. We hypothesise that PubHealth being a manually curated dataset leads to overall cleaner post-editing explanations, which annotators prefer. 

\subsection{Explanation Informativeness}~\label{sec:results:manual:info}
\textbf{Metrics.} We also perform a manual evaluation for veracity prediction. We ask annotators to provide a veracity label for a claim and an explanation where, same as for the evaluation of Explanation Quality, the explanations are either our system's input or output. The annotators provide a veracity label for three-way classification; true, false, and insufficient (see map to original labels for both datasets in Appendix \ref{sec:appendix}). We use 30 instances of explanation type and perform evaluation for both datasets with two annotators for each dataset and instance.

\textbf{Results.} 
For LIAR-PLUS, one annotator gave the correct label 80\% times for input and 67\% times for the output explanations. The second annotator chose the correct label 56\% times using output explanations and 44\% times using input explanations. However, both annotators found at least 16\% of explanations to be insufficient for veracity prediction (Table \ref{tab:results:humanevals_task2}). For PubHealth, both annotators found each explanation to be useful for the task. The first annotator chose the correct label 50\% \& 40\% of the times for the given input \& output explanations. The second annotator chose the correct label in 70\% of the cases for both explanations. This corroborates that for a clean dataset like PubHealth our explanations help for the task of veracity prediction.
\begin{table}[t]
\centering
\setlength{\tabcolsep}{3pt}
\fontsize{10}{9}\selectfont
\begin{tabular}{llrrrrrr} 
\toprule
\multirow{2}{*}{\textbf{Criterion}} & & \multicolumn{3}{c}{\bf LIAR-PLUS}  & \multicolumn{3}{c}{\bf PubHealth} \\
& \# & \bf Top-6 & \bf T-6+E-6+P & \bf Both & \bf Top-5 & \bf T-5+E-5+P & \bf Both \\
\midrule
\multirow{2}{*}{Coverage} & 1 & \textbf{42.5} & 0.0 & 57.5 & 27.5 & \textbf{60.0} & 12.5 \\
& 2 & \textbf{40.0} & 5.0  & 55.0 &  \textbf{22.5} & 20.0 & 57.5 \\
\midrule
\multirow{2}{*}{Non-redundancy} & 1 & 10.0 & \textbf{87.5}  & 2.5 &  10.0 & \textbf{82.5}  & 7.5 \\
& 2 & 7.5  & \textbf{10.0}  & 82.5 &  7.5 & \textbf{75.0} & 17.5 \\

\midrule
\multirow{2}{*}{Non-contradictory} & 1 & \textbf{32.5}  & 5.0 & 62.5 & 7.5 & \textbf{10.0} & 82.5 \\
& 2 &  \textbf{10.0}  & 7.5 & 82.5 & \textbf{20.0} & 15.0 & 65.0 \\
\midrule
\multirow{2}{*}{Fluency} & 1 & 40.0 & \textbf{57.5} & 2.5 & 35.0 & \textbf{52.5} & 12.5 \\
& 2 &  \textbf{77.5}  & 15.0 & 7.5 & 20.0 & \textbf{72.5} & 7.5 \\

\midrule
\multirow{2}{*}{Fluency} & 1 & \textbf{57.5} & 42.5 & 0.0 & 35.0 & \textbf{62.5} & 2.5 \\
& 2 &  \textbf{62.5} & 15.0 & 22.5 & 25.0  & \textbf{67.5} & 7.5 \\

\bottomrule
\end{tabular}
\caption{Manual annotation results of explanation quality with two annotators for both datasets. Each value indicates the relative proportion of when an annotator preferred a justification for a criterion. The preferred method, out of the input Top-N (Supervised) and the output of our method, Top-N+Edits-N+Para, is emboldened, Both indicates no preference. 
}
\label{tab:results:humanevals_task1}
\end{table}
\begin{table}[t]
\centering
\fontsize{10}{10}\selectfont
\begin{tabular}{llrrrrrr}
\toprule
\multirow{2}{*}{\bf \#} & \multirow{2}{*}{\bf Explanation Type} & \multicolumn{3}{c}{\bf LIAR-PLUS}  &  \multicolumn{3}{c}{\bf PubHealth} \\
 &  & \bf M & \bf NM & \bf I & \bf M & \bf NM & \bf I  \\
\midrule
1 & Top-N (Supervised) & 20 & 5 & 5 & 15 & 15 & 0 \\
1 & Top-N+Edits-6+Para & 14 & 7 & 9 & 12 & 18 & 0 \\
\midrule
2 & Top-N (Supervised) & 11 & 14 & 5 & 21 & 9 & 0 \\
2 & Top-N+Edits-5+Para & 13 & 10 & 7 & 21 & 9 & 0 \\
\bottomrule
\end{tabular}
\caption{Manual evaluation results for predicting a veracity label. \# refers to annotator number, M/NM refers to number of matches/non-matches between annotator and original labels, I refers to number of explanations that were found to be insufficient to predict a label.}
\label{tab:results:humanevals_task2}
\end{table}
\begin{table*}[t]
\small
\centering
\begin{tabular}{p{419pt}}
\toprule
\textbf{Top-5:} Heavily-armed Muslims shouting “Allahu Akbar” open fire \myul[green]{campers and hikers} in a park. A heavily armed group of Middle Eastern looking Muslim men was arrested \myul[green]{outside Los Angeles} after opening fire upon hikers and campers in a large State Park \myul[orange]{in the area}. There was no evidence found that a crime had been committed by any of the subjects \myul[orange]{who were detained and they were released}. Also, the police report described the men only as “\myul[green]{males},” not “Middle Eastern \myul[orange]{males}” or “Muslim \myul[orange]{males}.” The website that started this rumor was Superstation95, which is not a “superstation” at all but rather a repository of misinformation from Hal Turner, who in 2010 was sentenced to 33 months in prison for making death threats against three federal judges. No credible news reports made any mention of the “Allahu Akbar” claim, and no witnesses stated they had been “shot at” \myul[orange]{by the men while hiking or camping.}
 \\
\hdashline
\textbf{Top-5+Edits-5:} \myul[blue]{Heavily-armed Muslims} \textbf{\textcolor{green}{\myul[blue]{males}}} \myul[blue]{shouting} “Allahu Akbar” open fire in a park. A heavily armed group of Middle Eastern looking Muslim men was arrested after opening fire upon hikers and campers in a large State Park \textbf{\textcolor{green}{outside Los Angeles}}. There was no evidence found that a crime had been committed \myul[blue]{by any of the subjects on}
\textbf{\textcolor{green}{\myul[blue]{campers and hikers}}}. \myul[orange]{Also, the police report described the men only as ``,'' not ``Middle Eastern'' or ``Muslim.''} The website that started this rumor was Superstation95, which is not a “superstation” at all but rather a repository of misinformation from Hal Turner, who in 2010 was sentenced to 33 months in prison \myul[blue]{for making death threats against three federal judges}. \myul[blue]{No credible news reports made any mention of the ``Allahu Akbar'' claim, and no witnesses stated they had been ``shot at''.}
 \\
\hdashline
\textbf{Top-5+Edits-5+Para:} \textbf{\textcolor{blue}{Muslims shout}} "Allahu Akbar" open fire in a park. A heavily armed group of Middle Eastern looking Muslim men was arrested after opening fire on hikers and campers in a large State Park outside Los Angeles. There was no evidence that a crime had been committed by \textbf{\textcolor{blue}{any of the campers or hikers}}. The website that started this rumor was Superstation95, which is not a “superstation” at all but rather a repository of misinformation from Hal Turner, who in 2010 was sentenced to 33 months in prison. \textbf{\textcolor{blue}{There were no credible news reports that mentioned the Allahu Akbar claim, and no witnesses that said they had been shot at.}}\\
\hdashline
\textbf{Original Explanation:} Secondary reporting claiming that Muslim men fired upon hikers (and that the media covered it up) appeared on a site that had previously inaccurately claimed Illinois had applied Sharia law to driver’s licenses, that Target introduced “Sharia-compliant” checkout lanes, and that Muslims successfully banned Halloween at a New Jersey school. \\
\hdashline
\textbf{Claim:} The media covered up an incident in San Bernardino during which several Muslim men fired upon a number of Californian hikers. \textbf{Label:} False\\
\midrule
\textbf{Top-5}: The article claims the CDC might have to stop calling COVID-19 an epidemic because the death rate is becoming so low that it wouldn't meet the CDC's definition of epidemic. The latest CDC statement \myul[orange]{made public when the Facebook post was made said deaths attributed to COVID-19 decreased} from the previous week, but \myul[orange]{remained at the epidemic threshold, and were likely to increase}. Moreover death rates \myul[orange]{alone} do not define an epidemic. Amid news headlines that the United States set a daily record for the number of new coronavirus cases,\myul[orange]{ an article widely shared on Facebook made a contrarian claim}. The CDC page says: ``Epidemic refers to an increase, often sudden, in the number of cases of a disease above what is normally expected \myul[green]{in that population} in that area.'' \\
\hdashline
\textbf{Top-5+Edits-5:} \myul[blue]{The article claims} the CDC might have to stop calling COVID-19 an epidemic \textbf{\textcolor{green}{\myul[blue]{in that population}}} because the death rate is \myul[blue]{becoming} so low that it wouldn't meet \myul[blue]{the CDC's} definition of epidemic. \myul[blue]{The latest CDC statement an article from the previous week said deaths decreased, but.} \myul[blue]{Moreover,} death rates do not define an epidemic. \myul[orange]{Amid news headlines that the United States} \textbf{\textcolor{green}{\myul[orange]{on Facebook}}} \myul[orange]{set a daily record for the number of new coronavirus cases,} The CDC page \myul[blue]{ when the Facebook post was made} says: ``Epidemic refers to an increase, often sudden, in the number of cases of a disease attributed to COVID-19.'' \\
\hdashline
\textbf{Top-5+Edits-5+Para:}
\textbf{\textcolor{blue}{According to the article}}, the CDC might have to stop calling COVID-19 an epidemic because the death rate is so low that it wouldn't meet \textbf{\textcolor{blue}{their}} definition of an epidemic. \textbf{\textcolor{blue}{An article from the previous week said deaths decreased, but that's what the latest CDC statement says.}} Death rates do not define an epidemic. The CDC's page on Facebook says Epidemic refers to an increase, often sudden, in the number of cases of a disease attributed to COVID-19. \\
\hdashline
\textbf{Original Explanation:}
Despite a dip in death rates, which are expected to rise again, the federal Centers for Disease Control and Prevention still considers COVID-19 an epidemic. Death rates alone don’t determine whether an outbreak is an epidemic. \\
\hdashline 
\textbf{Claim:} The CDC may have to stop calling COVID-19 an ‘epidemic’ due to a remarkably low death rate. \textbf{Label:} False \\
\bottomrule
\end{tabular}
\caption{Example explanations -- extracted Top-5 RCs, the iterative editing, and the latter with paraphrasing on top, taken from the test split of PubHealth. Each color designates an edit operation -- \textbf{\textcolor{green}{reordering}}, \textbf{\textcolor{orange}{deletion}}, and \textbf{\textcolor{blue}{paraphrasing}}. The underlining designates the position in the text where the corresponding operation will be applied in the next step -- post-editing and paraphrasing.} 
\label{tab:qualexamples}
\end{table*}

\section{Discussion}~\label{sec:discussion}
Results from our automatic and manual evaluation suggest two main implications of applying our post-editing algorithm over extracted RCs. 
First, with the automatic ROUGE evaluation, we confirmed that the post-editing
preserves a large portion of important information that is contained in the gold explanation and is important for the fact check.
This was further supported by our manual evaluation of veracity predictions, where the post-edited explanations have been most useful for predicting the correct label. We conjecture the above indicates that our post-editing can be applied more generally for automated summarisation for knowledge-intensive tasks, such as fact checking and question answering, where the information needed for prediction has to be preserved. 

Second, with both the automatic and manual evaluation, we also corroborate that our proposed post-editing method improves several qualities of the generated explanations -- fluency, conciseness, and readability. The latter supports the usefulness of the length and fluency scores as well as the grammatical correction and the paraphrasing steps promoting these particular qualities of the generated explanations. Fluency, conciseness, and readability are important prerequisites for building trust in automated fact checking predictions especially for systems used in practice as \citet{thagard1989explanatory} find that people generally prefer simpler, more general explanations with fewer causes.  They can also contribute to reaching a broader audience when conveying the veracity of the claim. Conciseness and readability are also the downsides of current professional long and in-depth RCs, which some leading fact checking organisations, e.g., PolitiFact,\footnote{https://www.politifact.com/} have slowly started addressing by including short overview sections for the RCs.

Table~\ref{tab:qualexamples} further presents a case study from the PubHealth dataset. Overall, the initial extracted RC sentences are transformed to be more concise, fluent and human-readable by applying the iterative post-editing algorithm followed by paraphrasing. We can also see that compared to the original explanation, the post-edited explanations contain words that do not change the semantics of the explanation, but would not be scored as correct according to the ROUGE scores. For example, in the second instance, ``Death rates do not define an epidemic'' in the post-edited explanation and ``Death rates alone don’t determine whether an outbreak is an epidemic'' from the original explanation express the same meaning, but contain both paraphrases and filler words that would decrease the final ROUGE scores. Finally, compared to the original explanation, the post-edited explanations for both instances have preserved the information needed for the fact checking.




\section{Conclusion}~\label{sec:conclusion}
In this work, we present an unsupervised post-editing approach to improve extractive explanations for fact-checking. Our novel approach is based on an iterative edit-based algorithm and rephrasing-based post-processing. In our experiments on two fact checking benchmarking datasets, we observe, in both the manual and automatic evaluation, that our approaches generate fluent, coherent, and semantics-preserving explanations. 

\section*{Acknowledgments}

\begin{wrapfigure}{L}{0.10\columnwidth}
\vspace{-13pt}
\includegraphics[width=0.1\columnwidth]{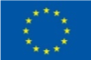}
\vspace{-25pt}
\end{wrapfigure}
Shailza Jolly was supported by the TU Kaiserslautern CS Ph.D. scholarship program, the BMBF project XAINES (Grant 01IW20005), a STSM grant from the COST project Multi3Generation (CA18231), and the NVIDIA AI Lab (NVAIL) program.
Pepa Atanasova has received funding from the European Union’s Horizon 2020 research and innovation programme under the Marie Skłodowska-Curie grant agreement No 801199. Isabelle Augenstein's research is further partially funded by a DFF Sapere Aude research leader grant.

\bibliographystyle{ACM-Reference-Format}
\bibliography{anthology.bib, custom.bib}


\begin{thebibliography}{40}


\ifx \showCODEN    \undefined \def \showCODEN     #1{\unskip}     \fi
\ifx \showDOI      \undefined \def \showDOI       #1{#1}\fi
\ifx \showISBNx    \undefined \def \showISBNx     #1{\unskip}     \fi
\ifx \showISBNxiii \undefined \def \showISBNxiii  #1{\unskip}     \fi
\ifx \showISSN     \undefined \def \showISSN      #1{\unskip}     \fi
\ifx \showLCCN     \undefined \def \showLCCN      #1{\unskip}     \fi
\ifx \shownote     \undefined \def \shownote      #1{#1}          \fi
\ifx \showarticletitle \undefined \def \showarticletitle #1{#1}   \fi
\ifx \showURL      \undefined \def \showURL       {\relax}        \fi
\providecommand\bibfield[2]{#2}
\providecommand\bibinfo[2]{#2}
\providecommand\natexlab[1]{#1}
\providecommand\showeprint[2][]{arXiv:#2}

\bibitem[\protect\citeauthoryear{Alhindi, Petridis, and Muresan}{Alhindi
  et~al\mbox{.}}{2018}]%
        {alhindi-etal-2018-evidence}
\bibfield{author}{\bibinfo{person}{Tariq Alhindi}, \bibinfo{person}{Savvas
  Petridis}, {and} \bibinfo{person}{Smaranda Muresan}.}
  \bibinfo{year}{2018}\natexlab{}.
\newblock \showarticletitle{Where is Your Evidence: Improving Fact-checking by
  Justification Modeling}. In \bibinfo{booktitle}{\emph{Proceedings of the
  First Workshop on Fact Extraction and {VER}ification ({FEVER})}}.
  \bibinfo{publisher}{Association for Computational Linguistics},
  \bibinfo{address}{Brussels, Belgium}, \bibinfo{pages}{85--90}.
\newblock
\urldef\tempurl%
\url{https://doi.org/10.18653/v1/W18-5513}
\showDOI{\tempurl}


\bibitem[\protect\citeauthoryear{Atanasova, Simonsen, Lioma, and
  Augenstein}{Atanasova et~al\mbox{.}}{2020a}]%
        {atanasova-etal-2020-diagnostic}
\bibfield{author}{\bibinfo{person}{Pepa Atanasova}, \bibinfo{person}{Jakob~Grue
  Simonsen}, \bibinfo{person}{Christina Lioma}, {and} \bibinfo{person}{Isabelle
  Augenstein}.} \bibinfo{year}{2020}\natexlab{a}.
\newblock \showarticletitle{A Diagnostic Study of Explainability Techniques for
  Text Classification}. In \bibinfo{booktitle}{\emph{Proceedings of the 2020
  Conference on Empirical Methods in Natural Language Processing (EMNLP)}}.
  \bibinfo{publisher}{Association for Computational Linguistics},
  \bibinfo{address}{Online}, \bibinfo{pages}{3256--3274}.
\newblock
\urldef\tempurl%
\url{https://doi.org/10.18653/v1/2020.emnlp-main.263}
\showDOI{\tempurl}


\bibitem[\protect\citeauthoryear{Atanasova, Simonsen, Lioma, and
  Augenstein}{Atanasova et~al\mbox{.}}{2020b}]%
        {atanasova-etal-2020-generating-fact}
\bibfield{author}{\bibinfo{person}{Pepa Atanasova}, \bibinfo{person}{Jakob~Grue
  Simonsen}, \bibinfo{person}{Christina Lioma}, {and} \bibinfo{person}{Isabelle
  Augenstein}.} \bibinfo{year}{2020}\natexlab{b}.
\newblock \showarticletitle{Generating Fact Checking Explanations}. In
  \bibinfo{booktitle}{\emph{Proceedings of the 58th Annual Meeting of the
  Association for Computational Linguistics}}. \bibinfo{publisher}{Association
  for Computational Linguistics}, \bibinfo{address}{Online},
  \bibinfo{pages}{7352--7364}.
\newblock
\urldef\tempurl%
\url{https://doi.org/10.18653/v1/2020.acl-main.656}
\showDOI{\tempurl}


\bibitem[\protect\citeauthoryear{Atanasova, Simonsen, Lioma, and
  Augenstein}{Atanasova et~al\mbox{.}}{2021}]%
        {atanasova2021diagnostics}
\bibfield{author}{\bibinfo{person}{Pepa Atanasova}, \bibinfo{person}{Jakob~Grue
  Simonsen}, \bibinfo{person}{Christina Lioma}, {and} \bibinfo{person}{Isabelle
  Augenstein}.} \bibinfo{year}{2021}\natexlab{}.
\newblock \showarticletitle{Diagnostics-Guided Explanation Generation}. In
  \bibinfo{booktitle}{\emph{Proceedings of the Thirty-Sixth AAAI Conference on
  Artificial Intelligence}} (Vancouver, Canada)
  \emph{(\bibinfo{series}{AAAI'21})}. \bibinfo{publisher}{AAAI Press},
  \bibinfo{numpages}{8}~pages.
\newblock


\bibitem[\protect\citeauthoryear{Atanasova, Wright, and Augenstein}{Atanasova
  et~al\mbox{.}}{2020c}]%
        {atanasova-etal-2020-generating}
\bibfield{author}{\bibinfo{person}{Pepa Atanasova}, \bibinfo{person}{Dustin
  Wright}, {and} \bibinfo{person}{Isabelle Augenstein}.}
  \bibinfo{year}{2020}\natexlab{c}.
\newblock \showarticletitle{Generating Label Cohesive and Well-Formed
  Adversarial Claims}. In \bibinfo{booktitle}{\emph{Proceedings of the 2020
  Conference on Empirical Methods in Natural Language Processing (EMNLP)}}.
  \bibinfo{publisher}{Association for Computational Linguistics},
  \bibinfo{address}{Online}, \bibinfo{pages}{3168--3177}.
\newblock
\urldef\tempurl%
\url{https://doi.org/10.18653/v1/2020.emnlp-main.256}
\showDOI{\tempurl}


\bibitem[\protect\citeauthoryear{Augenstein, Lioma, Wang, Chaves~Lima, Hansen,
  Hansen, and Simonsen}{Augenstein et~al\mbox{.}}{2019}]%
        {augenstein-etal-2019-multifc}
\bibfield{author}{\bibinfo{person}{Isabelle Augenstein},
  \bibinfo{person}{Christina Lioma}, \bibinfo{person}{Dongsheng Wang},
  \bibinfo{person}{Lucas Chaves~Lima}, \bibinfo{person}{Casper Hansen},
  \bibinfo{person}{Christian Hansen}, {and} \bibinfo{person}{Jakob~Grue
  Simonsen}.} \bibinfo{year}{2019}\natexlab{}.
\newblock \showarticletitle{{M}ulti{FC}: A Real-World Multi-Domain Dataset for
  Evidence-Based Fact Checking of Claims}. In
  \bibinfo{booktitle}{\emph{Proceedings of the 2019 Conference on Empirical
  Methods in Natural Language Processing and the 9th International Joint
  Conference on Natural Language Processing (EMNLP-IJCNLP)}}.
  \bibinfo{publisher}{Association for Computational Linguistics},
  \bibinfo{address}{Hong Kong, China}, \bibinfo{pages}{4685--4697}.
\newblock
\urldef\tempurl%
\url{https://doi.org/10.18653/v1/D19-1475}
\showDOI{\tempurl}


\bibitem[\protect\citeauthoryear{Beltagy, Lo, and Cohan}{Beltagy
  et~al\mbox{.}}{2019}]%
        {beltagy-etal-2019-scibert}
\bibfield{author}{\bibinfo{person}{Iz Beltagy}, \bibinfo{person}{Kyle Lo},
  {and} \bibinfo{person}{Arman Cohan}.} \bibinfo{year}{2019}\natexlab{}.
\newblock \showarticletitle{{S}ci{BERT}: A Pretrained Language Model for
  Scientific Text}. In \bibinfo{booktitle}{\emph{Proceedings of the 2019
  Conference on Empirical Methods in Natural Language Processing and the 9th
  International Joint Conference on Natural Language Processing
  (EMNLP-IJCNLP)}}. \bibinfo{publisher}{Association for Computational
  Linguistics}, \bibinfo{address}{Hong Kong, China},
  \bibinfo{pages}{3615--3620}.
\newblock
\urldef\tempurl%
\url{https://doi.org/10.18653/v1/D19-1371}
\showDOI{\tempurl}


\bibitem[\protect\citeauthoryear{Beltagy, Peters, and Cohan}{Beltagy
  et~al\mbox{.}}{2020}]%
        {Beltagy2020Longformer}
\bibfield{author}{\bibinfo{person}{Iz Beltagy}, \bibinfo{person}{Matthew~E.
  Peters}, {and} \bibinfo{person}{Arman Cohan}.}
  \bibinfo{year}{2020}\natexlab{}.
\newblock \bibinfo{title}{Longformer: The Long-Document Transformer}.
\newblock
\newblock
\showeprint[arxiv]{2004.05150}~[cs.CL]


\bibitem[\protect\citeauthoryear{Brown, Mann, Ryder, Subbiah, Kaplan, Dhariwal,
  Neelakantan, Shyam, Sastry, Askell, Agarwal, Herbert-Voss, Krueger, Henighan,
  Child, Ramesh, Ziegler, Wu, Winter, Hesse, Chen, Sigler, Litwin, Gray, Chess,
  Clark, Berner, McCandlish, Radford, Sutskever, and Amodei}{Brown
  et~al\mbox{.}}{2020}]%
        {brown2020language}
\bibfield{author}{\bibinfo{person}{Tom Brown}, \bibinfo{person}{Benjamin Mann},
  \bibinfo{person}{Nick Ryder}, \bibinfo{person}{Melanie Subbiah},
  \bibinfo{person}{Jared~D Kaplan}, \bibinfo{person}{Prafulla Dhariwal},
  \bibinfo{person}{Arvind Neelakantan}, \bibinfo{person}{Pranav Shyam},
  \bibinfo{person}{Girish Sastry}, \bibinfo{person}{Amanda Askell},
  \bibinfo{person}{Sandhini Agarwal}, \bibinfo{person}{Ariel Herbert-Voss},
  \bibinfo{person}{Gretchen Krueger}, \bibinfo{person}{Tom Henighan},
  \bibinfo{person}{Rewon Child}, \bibinfo{person}{Aditya Ramesh},
  \bibinfo{person}{Daniel Ziegler}, \bibinfo{person}{Jeffrey Wu},
  \bibinfo{person}{Clemens Winter}, \bibinfo{person}{Chris Hesse},
  \bibinfo{person}{Mark Chen}, \bibinfo{person}{Eric Sigler},
  \bibinfo{person}{Mateusz Litwin}, \bibinfo{person}{Scott Gray},
  \bibinfo{person}{Benjamin Chess}, \bibinfo{person}{Jack Clark},
  \bibinfo{person}{Christopher Berner}, \bibinfo{person}{Sam McCandlish},
  \bibinfo{person}{Alec Radford}, \bibinfo{person}{Ilya Sutskever}, {and}
  \bibinfo{person}{Dario Amodei}.} \bibinfo{year}{2020}\natexlab{}.
\newblock \showarticletitle{Language Models are Few-Shot Learners}. In
  \bibinfo{booktitle}{\emph{Advances in Neural Information Processing
  Systems}}, \bibfield{editor}{\bibinfo{person}{H.~Larochelle},
  \bibinfo{person}{M.~Ranzato}, \bibinfo{person}{R.~Hadsell},
  \bibinfo{person}{M.~F. Balcan}, {and} \bibinfo{person}{H.~Lin}} (Eds.),
  Vol.~\bibinfo{volume}{33}. \bibinfo{publisher}{Curran Associates, Inc.},
  \bibinfo{pages}{1877--1901}.
\newblock
\urldef\tempurl%
\url{https://proceedings.neurips.cc/paper/2020/file/1457c0d6bfcb4967418bfb8ac142f64a-Paper.pdf}
\showURL{%
\tempurl}


\bibitem[\protect\citeauthoryear{Camburu, Rockt\"{a}schel, Lukasiewicz, and
  Blunsom}{Camburu et~al\mbox{.}}{2018}]%
        {camburu2018snli}
\bibfield{author}{\bibinfo{person}{Oana-Maria Camburu}, \bibinfo{person}{Tim
  Rockt\"{a}schel}, \bibinfo{person}{Thomas Lukasiewicz}, {and}
  \bibinfo{person}{Phil Blunsom}.} \bibinfo{year}{2018}\natexlab{}.
\newblock \showarticletitle{{e-SNLI: Natural Language Inference with Natural
  Language Explanations}}.
\newblock In \bibinfo{booktitle}{\emph{Advances in Neural Information
  Processing Systems 31}}, \bibfield{editor}{\bibinfo{person}{S.~Bengio},
  \bibinfo{person}{H.~Wallach}, \bibinfo{person}{H.~Larochelle},
  \bibinfo{person}{K.~Grauman}, \bibinfo{person}{N.~Cesa-Bianchi}, {and}
  \bibinfo{person}{R.~Garnett}} (Eds.). \bibinfo{publisher}{Curran Associates,
  Inc.}, \bibinfo{pages}{9539--9549}.
\newblock
\urldef\tempurl%
\url{http://papers.nips.cc/paper/8163-e-snli-natural-language-inference-with-natural-language-explanations.pdf}
\showURL{%
\tempurl}


\bibitem[\protect\citeauthoryear{Chen and Bansal}{Chen and Bansal}{2018}]%
        {chen-bansal-2018-fast}
\bibfield{author}{\bibinfo{person}{Yen-Chun Chen} {and} \bibinfo{person}{Mohit
  Bansal}.} \bibinfo{year}{2018}\natexlab{}.
\newblock \showarticletitle{Fast Abstractive Summarization with
  Reinforce-Selected Sentence Rewriting}. In
  \bibinfo{booktitle}{\emph{Proceedings of the 56th Annual Meeting of the
  Association for Computational Linguistics (Volume 1: Long Papers)}}.
  \bibinfo{publisher}{Association for Computational Linguistics},
  \bibinfo{address}{Melbourne, Australia}, \bibinfo{pages}{675--686}.
\newblock
\urldef\tempurl%
\url{https://doi.org/10.18653/v1/P18-1063}
\showDOI{\tempurl}


\bibitem[\protect\citeauthoryear{DeYoung, Jain, Rajani, Lehman, Xiong, Socher,
  and Wallace}{DeYoung et~al\mbox{.}}{2020}]%
        {deyoung-etal-2020-eraser}
\bibfield{author}{\bibinfo{person}{Jay DeYoung}, \bibinfo{person}{Sarthak
  Jain}, \bibinfo{person}{Nazneen~Fatema Rajani}, \bibinfo{person}{Eric
  Lehman}, \bibinfo{person}{Caiming Xiong}, \bibinfo{person}{Richard Socher},
  {and} \bibinfo{person}{Byron~C. Wallace}.} \bibinfo{year}{2020}\natexlab{}.
\newblock \showarticletitle{{ERASER}: {A} Benchmark to Evaluate Rationalized
  {NLP} Models}. In \bibinfo{booktitle}{\emph{Proceedings of the 58th Annual
  Meeting of the Association for Computational Linguistics}}.
  \bibinfo{publisher}{Association for Computational Linguistics},
  \bibinfo{address}{Online}, \bibinfo{pages}{4443--4458}.
\newblock
\urldef\tempurl%
\url{https://doi.org/10.18653/v1/2020.acl-main.408}
\showDOI{\tempurl}


\bibitem[\protect\citeauthoryear{Hinton}{Hinton}{2002}]%
        {hinton2002training}
\bibfield{author}{\bibinfo{person}{Geoffrey~E Hinton}.}
  \bibinfo{year}{2002}\natexlab{}.
\newblock \showarticletitle{Training products of experts by minimizing
  contrastive divergence}.
\newblock \bibinfo{journal}{\emph{Neural computation}} \bibinfo{volume}{14},
  \bibinfo{number}{8} (\bibinfo{year}{2002}), \bibinfo{pages}{1771--1800}.
\newblock


\bibitem[\protect\citeauthoryear{Kincaid, Fishburne~Jr, Rogers, and
  Chissom}{Kincaid et~al\mbox{.}}{1975}]%
        {kincaid1975derivation}
\bibfield{author}{\bibinfo{person}{J~Peter Kincaid}, \bibinfo{person}{Robert~P
  Fishburne~Jr}, \bibinfo{person}{Richard~L Rogers}, {and}
  \bibinfo{person}{Brad~S Chissom}.} \bibinfo{year}{1975}\natexlab{}.
\newblock \bibinfo{booktitle}{\emph{Derivation of new readability formulas
  (automated readability index, fog count and flesch reading ease formula) for
  navy enlisted personnel}}.
\newblock \bibinfo{type}{{T}echnical {R}eport}. \bibinfo{institution}{Naval
  Technical Training Command Millington TN Research Branch}.
\newblock


\bibitem[\protect\citeauthoryear{Kindermans, Hooker, Adebayo, Alber,
  Sch{\"u}tt, D{\"a}hne, Erhan, and Kim}{Kindermans et~al\mbox{.}}{2019}]%
        {kindermans2019reliability}
\bibfield{author}{\bibinfo{person}{Pieter-Jan Kindermans},
  \bibinfo{person}{Sara Hooker}, \bibinfo{person}{Julius Adebayo},
  \bibinfo{person}{Maximilian Alber}, \bibinfo{person}{Kristof~T Sch{\"u}tt},
  \bibinfo{person}{Sven D{\"a}hne}, \bibinfo{person}{Dumitru Erhan}, {and}
  \bibinfo{person}{Been Kim}.} \bibinfo{year}{2019}\natexlab{}.
\newblock \showarticletitle{The (un) reliability of saliency methods}.
\newblock In \bibinfo{booktitle}{\emph{Explainable AI: Interpreting, Explaining
  and Visualizing Deep Learning}}. \bibinfo{publisher}{Springer},
  \bibinfo{pages}{267--280}.
\newblock


\bibitem[\protect\citeauthoryear{Kotonya and Toni}{Kotonya and Toni}{2020a}]%
        {kotonya-toni-2020-explainable-automated}
\bibfield{author}{\bibinfo{person}{Neema Kotonya} {and}
  \bibinfo{person}{Francesca Toni}.} \bibinfo{year}{2020}\natexlab{a}.
\newblock \showarticletitle{Explainable Automated Fact-Checking: A Survey}. In
  \bibinfo{booktitle}{\emph{Proceedings of the 28th International Conference on
  Computational Linguistics}}. \bibinfo{publisher}{International Committee on
  Computational Linguistics}, \bibinfo{address}{Barcelona, Spain (Online)},
  \bibinfo{pages}{5430--5443}.
\newblock
\urldef\tempurl%
\url{https://doi.org/10.18653/v1/2020.coling-main.474}
\showDOI{\tempurl}


\bibitem[\protect\citeauthoryear{Kotonya and Toni}{Kotonya and Toni}{2020b}]%
        {kotonya-toni-2020-explainable}
\bibfield{author}{\bibinfo{person}{Neema Kotonya} {and}
  \bibinfo{person}{Francesca Toni}.} \bibinfo{year}{2020}\natexlab{b}.
\newblock \showarticletitle{Explainable Automated Fact-Checking for Public
  Health Claims}. In \bibinfo{booktitle}{\emph{Proceedings of the 2020
  Conference on Empirical Methods in Natural Language Processing (EMNLP)}}.
  \bibinfo{publisher}{Association for Computational Linguistics},
  \bibinfo{address}{Online}, \bibinfo{pages}{7740--7754}.
\newblock
\urldef\tempurl%
\url{https://doi.org/10.18653/v1/2020.emnlp-main.623}
\showDOI{\tempurl}


\bibitem[\protect\citeauthoryear{Kumar, Mou, Golab, and Vechtomova}{Kumar
  et~al\mbox{.}}{2020}]%
        {kumar-etal-2020-iterative}
\bibfield{author}{\bibinfo{person}{Dhruv Kumar}, \bibinfo{person}{Lili Mou},
  \bibinfo{person}{Lukasz Golab}, {and} \bibinfo{person}{Olga Vechtomova}.}
  \bibinfo{year}{2020}\natexlab{}.
\newblock \showarticletitle{Iterative Edit-Based Unsupervised Sentence
  Simplification}. In \bibinfo{booktitle}{\emph{Proceedings of the 58th Annual
  Meeting of the Association for Computational Linguistics}}.
  \bibinfo{publisher}{Association for Computational Linguistics},
  \bibinfo{address}{Online}, \bibinfo{pages}{7918--7928}.
\newblock
\urldef\tempurl%
\url{https://doi.org/10.18653/v1/2020.acl-main.707}
\showDOI{\tempurl}


\bibitem[\protect\citeauthoryear{Li, Li, Mou, Jiang, Lyu, and King}{Li
  et~al\mbox{.}}{2020}]%
        {NEURIPS2020_7a677bb4}
\bibfield{author}{\bibinfo{person}{Jingjing Li}, \bibinfo{person}{Zichao Li},
  \bibinfo{person}{Lili Mou}, \bibinfo{person}{Xin Jiang},
  \bibinfo{person}{Michael Lyu}, {and} \bibinfo{person}{Irwin King}.}
  \bibinfo{year}{2020}\natexlab{}.
\newblock \showarticletitle{Unsupervised Text Generation by Learning from
  Search}. In \bibinfo{booktitle}{\emph{Advances in Neural Information
  Processing Systems}}, \bibfield{editor}{\bibinfo{person}{H.~Larochelle},
  \bibinfo{person}{M.~Ranzato}, \bibinfo{person}{R.~Hadsell},
  \bibinfo{person}{M.~F. Balcan}, {and} \bibinfo{person}{H.~Lin}} (Eds.),
  Vol.~\bibinfo{volume}{33}. \bibinfo{publisher}{Curran Associates, Inc.},
  \bibinfo{address}{Online}, \bibinfo{pages}{10820--10831}.
\newblock
\urldef\tempurl%
\url{https://proceedings.neurips.cc/paper/2020/file/7a677bb4477ae2dd371add568dd19e23-Paper.pdf}
\showURL{%
\tempurl}


\bibitem[\protect\citeauthoryear{Liu, Mou, Meng, Zhou, Zhou, and Song}{Liu
  et~al\mbox{.}}{2020}]%
        {liu-etal-2020-unsupervised}
\bibfield{author}{\bibinfo{person}{Xianggen Liu}, \bibinfo{person}{Lili Mou},
  \bibinfo{person}{Fandong Meng}, \bibinfo{person}{Hao Zhou},
  \bibinfo{person}{Jie Zhou}, {and} \bibinfo{person}{Sen Song}.}
  \bibinfo{year}{2020}\natexlab{}.
\newblock \showarticletitle{Unsupervised Paraphrasing by Simulated Annealing}.
  In \bibinfo{booktitle}{\emph{Proceedings of the 58th Annual Meeting of the
  Association for Computational Linguistics}}. \bibinfo{publisher}{Association
  for Computational Linguistics}, \bibinfo{address}{Online},
  \bibinfo{pages}{302--312}.
\newblock
\urldef\tempurl%
\url{https://doi.org/10.18653/v1/2020.acl-main.28}
\showDOI{\tempurl}


\bibitem[\protect\citeauthoryear{Liu and Lapata}{Liu and Lapata}{2019}]%
        {liu-lapata-2019-text}
\bibfield{author}{\bibinfo{person}{Yang Liu} {and} \bibinfo{person}{Mirella
  Lapata}.} \bibinfo{year}{2019}\natexlab{}.
\newblock \showarticletitle{Text Summarization with Pretrained Encoders}. In
  \bibinfo{booktitle}{\emph{Proceedings of the 2019 Conference on Empirical
  Methods in Natural Language Processing and the 9th International Joint
  Conference on Natural Language Processing (EMNLP-IJCNLP)}}.
  \bibinfo{publisher}{Association for Computational Linguistics},
  \bibinfo{address}{Hong Kong, China}, \bibinfo{pages}{3730--3740}.
\newblock
\urldef\tempurl%
\url{https://doi.org/10.18653/v1/D19-1387}
\showDOI{\tempurl}


\bibitem[\protect\citeauthoryear{Liu, Ott, Goyal, Du, Joshi, Chen, Levy, Lewis,
  Zettlemoyer, and Stoyanov}{Liu et~al\mbox{.}}{2019}]%
        {liu2019roberta}
\bibfield{author}{\bibinfo{person}{Yinhan Liu}, \bibinfo{person}{Myle Ott},
  \bibinfo{person}{Naman Goyal}, \bibinfo{person}{Jingfei Du},
  \bibinfo{person}{Mandar Joshi}, \bibinfo{person}{Danqi Chen},
  \bibinfo{person}{Omer Levy}, \bibinfo{person}{Mike Lewis},
  \bibinfo{person}{Luke Zettlemoyer}, {and} \bibinfo{person}{Veselin
  Stoyanov}.} \bibinfo{year}{2019}\natexlab{}.
\newblock \bibinfo{title}{RoBERTa: A Robustly Optimized BERT Pretraining
  Approach}.
\newblock
\newblock
\showeprint[arxiv]{1907.11692}~[cs.CL]


\bibitem[\protect\citeauthoryear{Lu and Li}{Lu and Li}{2020}]%
        {lu-li-2020-gcan}
\bibfield{author}{\bibinfo{person}{Yi-Ju Lu} {and} \bibinfo{person}{Cheng-Te
  Li}.} \bibinfo{year}{2020}\natexlab{}.
\newblock \showarticletitle{{GCAN}: Graph-aware Co-Attention Networks for
  Explainable Fake News Detection on Social Media}. In
  \bibinfo{booktitle}{\emph{Proceedings of the 58th Annual Meeting of the
  Association for Computational Linguistics}}. \bibinfo{publisher}{Association
  for Computational Linguistics}, \bibinfo{address}{Online},
  \bibinfo{pages}{505--514}.
\newblock
\urldef\tempurl%
\url{https://doi.org/10.18653/v1/2020.acl-main.48}
\showDOI{\tempurl}


\bibitem[\protect\citeauthoryear{Manning, Surdeanu, Bauer, Finkel, Bethard, and
  McClosky}{Manning et~al\mbox{.}}{2014}]%
        {manning-etal-2014-stanford}
\bibfield{author}{\bibinfo{person}{Christopher Manning}, \bibinfo{person}{Mihai
  Surdeanu}, \bibinfo{person}{John Bauer}, \bibinfo{person}{Jenny Finkel},
  \bibinfo{person}{Steven Bethard}, {and} \bibinfo{person}{David McClosky}.}
  \bibinfo{year}{2014}\natexlab{}.
\newblock \showarticletitle{The {S}tanford {C}ore{NLP} Natural Language
  Processing Toolkit}. In \bibinfo{booktitle}{\emph{Proceedings of 52nd Annual
  Meeting of the Association for Computational Linguistics: System
  Demonstrations}}. \bibinfo{publisher}{Association for Computational
  Linguistics}, \bibinfo{address}{Baltimore, Maryland},
  \bibinfo{pages}{55--60}.
\newblock
\urldef\tempurl%
\url{https://doi.org/10.3115/v1/P14-5010}
\showDOI{\tempurl}


\bibitem[\protect\citeauthoryear{Mishra, Gupta, and Leippold}{Mishra
  et~al\mbox{.}}{2020}]%
        {mishra-etal-2020-generating}
\bibfield{author}{\bibinfo{person}{Rahul Mishra}, \bibinfo{person}{Dhruv
  Gupta}, {and} \bibinfo{person}{Markus Leippold}.}
  \bibinfo{year}{2020}\natexlab{}.
\newblock \showarticletitle{Generating Fact Checking Summaries for Web Claims}.
  In \bibinfo{booktitle}{\emph{Proceedings of the Sixth Workshop on Noisy
  User-generated Text (W-NUT 2020)}}. \bibinfo{publisher}{Association for
  Computational Linguistics}, \bibinfo{address}{Online},
  \bibinfo{pages}{81--90}.
\newblock
\urldef\tempurl%
\url{https://doi.org/10.18653/v1/2020.wnut-1.12}
\showDOI{\tempurl}


\bibitem[\protect\citeauthoryear{Nallapati, Zhai, and Zhou}{Nallapati
  et~al\mbox{.}}{2017}]%
        {nallapati2017summarunner}
\bibfield{author}{\bibinfo{person}{Ramesh Nallapati}, \bibinfo{person}{Feifei
  Zhai}, {and} \bibinfo{person}{Bowen Zhou}.} \bibinfo{year}{2017}\natexlab{}.
\newblock \showarticletitle{SummaRuNNer: A Recurrent Neural Network Based
  Sequence Model for Extractive Summarization of Documents}. In
  \bibinfo{booktitle}{\emph{Proceedings of the Thirty-First AAAI Conference on
  Artificial Intelligence}} (San Francisco, California, USA)
  \emph{(\bibinfo{series}{AAAI'17})}. \bibinfo{publisher}{AAAI Press},
  \bibinfo{pages}{3075–3081}.
\newblock


\bibitem[\protect\citeauthoryear{Powers, Sumner, and Kearl}{Powers
  et~al\mbox{.}}{1958}]%
        {powers1958recalculation}
\bibfield{author}{\bibinfo{person}{Richard~D Powers},
  \bibinfo{person}{William~A Sumner}, {and} \bibinfo{person}{Bryant~E Kearl}.}
  \bibinfo{year}{1958}\natexlab{}.
\newblock \showarticletitle{A recalculation of four adult readability
  formulas.}
\newblock \bibinfo{journal}{\emph{Journal of Educational Psychology}}
  \bibinfo{volume}{49}, \bibinfo{number}{2} (\bibinfo{year}{1958}),
  \bibinfo{pages}{99}.
\newblock


\bibitem[\protect\citeauthoryear{Radford, Wu, Child, Luan, Amodei, and
  Sutskever}{Radford et~al\mbox{.}}{2019}]%
        {radford2019language}
\bibfield{author}{\bibinfo{person}{Alec Radford}, \bibinfo{person}{Jeffrey Wu},
  \bibinfo{person}{Rewon Child}, \bibinfo{person}{David Luan},
  \bibinfo{person}{Dario Amodei}, {and} \bibinfo{person}{Ilya Sutskever}.}
  \bibinfo{year}{2019}\natexlab{}.
\newblock \showarticletitle{Language models are unsupervised multitask
  learners}.
\newblock \bibinfo{journal}{\emph{OpenAI blog}} \bibinfo{volume}{1},
  \bibinfo{number}{8} (\bibinfo{year}{2019}), \bibinfo{pages}{9}.
\newblock


\bibitem[\protect\citeauthoryear{Reimers and Gurevych}{Reimers and
  Gurevych}{2019}]%
        {reimers-gurevych-2019-sentence}
\bibfield{author}{\bibinfo{person}{Nils Reimers} {and} \bibinfo{person}{Iryna
  Gurevych}.} \bibinfo{year}{2019}\natexlab{}.
\newblock \showarticletitle{Sentence-{BERT}: Sentence Embeddings using
  {S}iamese {BERT}-Networks}. In \bibinfo{booktitle}{\emph{Proceedings of the
  2019 Conference on Empirical Methods in Natural Language Processing and the
  9th International Joint Conference on Natural Language Processing
  (EMNLP-IJCNLP)}}. \bibinfo{publisher}{Association for Computational
  Linguistics}, \bibinfo{address}{Hong Kong, China},
  \bibinfo{pages}{3982--3992}.
\newblock
\urldef\tempurl%
\url{https://doi.org/10.18653/v1/D19-1410}
\showDOI{\tempurl}


\bibitem[\protect\citeauthoryear{Rose, Engel, Cramer, and Cowley}{Rose
  et~al\mbox{.}}{2010}]%
        {rose2010automatic}
\bibfield{author}{\bibinfo{person}{Stuart Rose}, \bibinfo{person}{Dave Engel},
  \bibinfo{person}{Nick Cramer}, {and} \bibinfo{person}{Wendy Cowley}.}
  \bibinfo{year}{2010}\natexlab{}.
\newblock \showarticletitle{Automatic keyword extraction from individual
  documents}.
\newblock \bibinfo{journal}{\emph{Text mining: applications and theory}}
  \bibinfo{volume}{1} (\bibinfo{year}{2010}), \bibinfo{pages}{1--20}.
\newblock


\bibitem[\protect\citeauthoryear{Sanh, Debut, Chaumond, and Wolf}{Sanh
  et~al\mbox{.}}{2020}]%
        {Sanh2019DistilBERTAD}
\bibfield{author}{\bibinfo{person}{Victor Sanh}, \bibinfo{person}{Lysandre
  Debut}, \bibinfo{person}{Julien Chaumond}, {and} \bibinfo{person}{Thomas
  Wolf}.} \bibinfo{year}{2020}\natexlab{}.
\newblock \bibinfo{title}{DistilBERT, a distilled version of BERT: smaller,
  faster, cheaper and lighter}.
\newblock
\newblock
\showeprint[arxiv]{1910.01108}~[cs.CL]


\bibitem[\protect\citeauthoryear{Schumann, Mou, Lu, Vechtomova, and
  Markert}{Schumann et~al\mbox{.}}{2020}]%
        {schumann-etal-2020-discrete}
\bibfield{author}{\bibinfo{person}{Raphael Schumann}, \bibinfo{person}{Lili
  Mou}, \bibinfo{person}{Yao Lu}, \bibinfo{person}{Olga Vechtomova}, {and}
  \bibinfo{person}{Katja Markert}.} \bibinfo{year}{2020}\natexlab{}.
\newblock \showarticletitle{Discrete Optimization for Unsupervised Sentence
  Summarization with Word-Level Extraction}. In
  \bibinfo{booktitle}{\emph{Proceedings of the 58th Annual Meeting of the
  Association for Computational Linguistics}}. \bibinfo{publisher}{Association
  for Computational Linguistics}, \bibinfo{address}{Online},
  \bibinfo{pages}{5032--5042}.
\newblock
\urldef\tempurl%
\url{https://doi.org/10.18653/v1/2020.acl-main.452}
\showDOI{\tempurl}


\bibitem[\protect\citeauthoryear{Simonyan, Vedaldi, and Zisserman}{Simonyan
  et~al\mbox{.}}{2014}]%
        {Simonyan2014DeepIC}
\bibfield{author}{\bibinfo{person}{K. Simonyan}, \bibinfo{person}{A. Vedaldi},
  {and} \bibinfo{person}{Andrew Zisserman}.} \bibinfo{year}{2014}\natexlab{}.
\newblock \showarticletitle{Deep Inside Convolutional Networks: Visualising
  Image Classification Models and Saliency Maps}.
\newblock \bibinfo{journal}{\emph{CoRR}}  \bibinfo{volume}{abs/1312.6034}
  (\bibinfo{year}{2014}).
\newblock


\bibitem[\protect\citeauthoryear{Stammbach and Ash}{Stammbach and Ash}{2020}]%
        {stammbach2020fever}
\bibfield{author}{\bibinfo{person}{Dominik Stammbach} {and}
  \bibinfo{person}{Elliott Ash}.} \bibinfo{year}{2020}\natexlab{}.
\newblock \showarticletitle{e-FEVER: Explanations and Summaries for Automated
  Fact Checking}. In \bibinfo{booktitle}{\emph{Proceedings of the 2020 Truth
  and Trust Online Conference (TTO 2020)}}. Hacks Hackers, \bibinfo{pages}{32}.
\newblock


\bibitem[\protect\citeauthoryear{Tan, Wan, and Xiao}{Tan et~al\mbox{.}}{2017}]%
        {tan-etal-2017-abstractive}
\bibfield{author}{\bibinfo{person}{Jiwei Tan}, \bibinfo{person}{Xiaojun Wan},
  {and} \bibinfo{person}{Jianguo Xiao}.} \bibinfo{year}{2017}\natexlab{}.
\newblock \showarticletitle{Abstractive Document Summarization with a
  Graph-Based Attentional Neural Model}. In
  \bibinfo{booktitle}{\emph{Proceedings of the 55th Annual Meeting of the
  Association for Computational Linguistics (Volume 1: Long Papers)}}.
  \bibinfo{publisher}{Association for Computational Linguistics},
  \bibinfo{address}{Vancouver, Canada}, \bibinfo{pages}{1171--1181}.
\newblock
\urldef\tempurl%
\url{https://doi.org/10.18653/v1/P17-1108}
\showDOI{\tempurl}


\bibitem[\protect\citeauthoryear{Thagard}{Thagard}{1989}]%
        {thagard1989explanatory}
\bibfield{author}{\bibinfo{person}{Paul Thagard}.}
  \bibinfo{year}{1989}\natexlab{}.
\newblock \showarticletitle{Explanatory coherence}.
\newblock \bibinfo{journal}{\emph{Behavioral and brain sciences}}
  \bibinfo{volume}{12}, \bibinfo{number}{3} (\bibinfo{year}{1989}),
  \bibinfo{pages}{435--467}.
\newblock


\bibitem[\protect\citeauthoryear{Thorne, Vlachos, Christodoulopoulos, and
  Mittal}{Thorne et~al\mbox{.}}{2018}]%
        {thorne-etal-2018-fever}
\bibfield{author}{\bibinfo{person}{James Thorne}, \bibinfo{person}{Andreas
  Vlachos}, \bibinfo{person}{Christos Christodoulopoulos}, {and}
  \bibinfo{person}{Arpit Mittal}.} \bibinfo{year}{2018}\natexlab{}.
\newblock \showarticletitle{{FEVER}: a Large-scale Dataset for Fact Extraction
  and {VER}ification}. In \bibinfo{booktitle}{\emph{Proceedings of the 2018
  Conference of the North {A}merican Chapter of the Association for
  Computational Linguistics: Human Language Technologies, Volume 1 (Long
  Papers)}}. \bibinfo{publisher}{Association for Computational Linguistics},
  \bibinfo{address}{New Orleans, Louisiana}, \bibinfo{pages}{809--819}.
\newblock
\urldef\tempurl%
\url{https://doi.org/10.18653/v1/N18-1074}
\showDOI{\tempurl}


\bibitem[\protect\citeauthoryear{Wang}{Wang}{2017}]%
        {wang-2017-liar}
\bibfield{author}{\bibinfo{person}{William~Yang Wang}.}
  \bibinfo{year}{2017}\natexlab{}.
\newblock \showarticletitle{{``}Liar, Liar Pants on Fire{''}: A New Benchmark
  Dataset for Fake News Detection}. In \bibinfo{booktitle}{\emph{Proceedings of
  the 55th Annual Meeting of the Association for Computational Linguistics
  (Volume 2: Short Papers)}}. \bibinfo{publisher}{Association for Computational
  Linguistics}, \bibinfo{address}{Vancouver, Canada},
  \bibinfo{pages}{422--426}.
\newblock
\urldef\tempurl%
\url{https://doi.org/10.18653/v1/P17-2067}
\showDOI{\tempurl}


\bibitem[\protect\citeauthoryear{Wu, Rao, Zhao, Liang, and Nazir}{Wu
  et~al\mbox{.}}{2020}]%
        {wu-etal-2020-dtca}
\bibfield{author}{\bibinfo{person}{Lianwei Wu}, \bibinfo{person}{Yuan Rao},
  \bibinfo{person}{Yongqiang Zhao}, \bibinfo{person}{Hao Liang}, {and}
  \bibinfo{person}{Ambreen Nazir}.} \bibinfo{year}{2020}\natexlab{}.
\newblock \showarticletitle{{DTCA}: Decision Tree-based Co-Attention Networks
  for Explainable Claim Verification}. In \bibinfo{booktitle}{\emph{Proceedings
  of the 58th Annual Meeting of the Association for Computational
  Linguistics}}. \bibinfo{publisher}{Association for Computational
  Linguistics}, \bibinfo{address}{Online}, \bibinfo{pages}{1024--1035}.
\newblock
\urldef\tempurl%
\url{https://doi.org/10.18653/v1/2020.acl-main.97}
\showDOI{\tempurl}


\bibitem[\protect\citeauthoryear{Zhang, Zhao, Saleh, and Liu}{Zhang
  et~al\mbox{.}}{2020}]%
        {zhang2020pegasus}
\bibfield{author}{\bibinfo{person}{Jingqing Zhang}, \bibinfo{person}{Yao Zhao},
  \bibinfo{person}{Mohammad Saleh}, {and} \bibinfo{person}{Peter Liu}.}
  \bibinfo{year}{2020}\natexlab{}.
\newblock \showarticletitle{Pegasus: Pre-training with extracted gap-sentences
  for abstractive summarization}. In \bibinfo{booktitle}{\emph{International
  Conference on Machine Learning}}. PMLR, \bibinfo{pages}{11328--11339}.
\newblock


\end{thebibliography}

\appendix
\label{sec:appendix}

\section{Manual Evaluation}


As explained in the Section \ref{sec:results:manual} of the main paper, we mapped user inputs (TRUE/FALSE) for task two to the original labels for each dataset. For Liar, we map "true", "mostly-true", "half-true" to TRUE and "false", "pants-fire", and "barely-true" to FALSE. In the PubHealth dataset, we map "true" to TRUE, "false" to FALSE.
The "insufficient" label is mapped to UNPROVEN. This way, once the mapping is done, we then compute the number of matches and non-matches to get an overall accuracy for this subset.

We appointed annotators with a university-level education in English. 

\section{Automatic Evaluation}
In Table~\ref{tab:results:rouge:val} and Table~\ref{tab:results:readability:val}, we provide results over the validation split of the datasets for the ROUGE and readability automatic evaluation. We additionally provide ablation results for components of our approach. First, applying Pegasus directly on the extracted sentences preserves a slightly larger amount of information when compared to applying Pegasus on top of the iterative editing approach -- up to 0.96 ROUGE-L scores, but the readability scores are still lower -- up to 4.28 Flesch Reading Ease points. We also show results of the two parts included in the Edits step -- the iterative editing and the grammar correction. We find that the grammar correction improves the ROUGE scores with up to 8 ROUGE-L score points and up to 8 Flesch Reading Ease points.


\section{Experimental Setup}
~\label{sec:experiments}

\subsection{Selection of Ruling Comments}
For the supervised selection of RCs, as described in Section~\ref{sec:method:selection}, we follow the implementation of the multi-task model of \citet{atanasova-etal-2020-generating-fact}. For LIAR-PLUS, we don't conduct fine-tuning as the model is already optimised for the dataset. For PubHealth, we change the base model to SciBERT, as the claims in PubHealth are from the health domain and previous work~\cite{kotonya-toni-2020-explainable} has shown that SciBERT outperforms BERTs for the domain. In Table~\ref{tab:finetune:ph:sup}, we show the results for the fine-tuning we performed over the multi-task architecture with a grid-search over the maximum length limit of the text and the weight for the positive sentences in the explanation extraction training objective. We finally select and use explanations generated with the multi-task model with a maximum text length of 1700, and a positive sentence weight of 5.

For the unsupervised selection of explanation sentences, we employ a Longformer model. We construct the Longformer model with BERT as a base architecture and conduct 2000 additional fine-tuning steps for the newly added cross-attention weights to be optimised. We then train models for both datasets supervised by veracity prediction. The most salient sentences are selected as the sentences that have the highest sum of token saliencies. 

Finally, we remove long sentences and questions from the RCs, where the ROUGE score changes after filtering are illustrated in Table~\ref{tab:finetune:sentence-scores}, which results in the Top-N sentences, that are used as input for the post-editing method.

These experiments were run on a single NVIDIA TitanRTX GPU with 24GB memory and 4 Intel Xeon Silver 4110 CPUs. Model training took $\sim 3$ hours.

\subsection{Iterative Based Algorithm}
We used the validation split of LIAR-PLUS to select the best hyperparameters for both datasets. We use the weight of 1.5, 1.2, 1.4, 0.95 for $\alpha$, $\eta$, $\gamma$, $\delta$ and 1.0 for $\beta$ in our scoring function. We set the thresholds as 0.94 for reordering, 0.97 for deletion, and 1.10 for insertion.  We keep all models -- GPT-2, RoBERTa, and Pegasus, fixed and do not finetune them on any in-house dataset. We run our search algorithm on a single V100-32 GB GPU for 220 steps, which takes around 13 hours for each split for both datasets. 

\section{Novelty and Copy Rate}
Table~\ref{tab:results:novelty} presents additional statistics for the generated explanations from the test sets of both datasets. First, we compute how many of the words from the input Top-N RCss are preserved in the final explanation. We find that with the final step of the post-editing process, up to 8\% of the tokens from the RCs are not found in the final explanation. On the other hand, our post-editing approach generates up to 10\% novel words that are not previously found in the RCs. This could explain the lower results for the ROUGE scores, which account only for exact token overlaps. Finally, while ROUGE scores are recall-oriented, i.e., they compute how many of the words in the gold explanation can be found in the candidate one, we compute a precision-oriented statistic of the words in the candidate that can be found in the gold explanation. Surprisingly, while ROUGE scores of our generated explanations decrease after post-processing, the reverse score increases, pointing to improvements in the precision-oriented overlap with our method.

In addition, in LIAR/PubHealth, the average summary length is 136/142 tokens for the extracted RCs, 89/86 for the gold justifications, 118.7/117.3 after iterative editing, and 98.5/94.7 after paraphrasing.

\begin{table*}[t]
\centering
\fontsize{10}{10}\selectfont
\begin{tabular}{lrrrrrr}
\toprule
& \multicolumn{3}{c}{\bf Validation} & \multicolumn{3}{c}{\bf Test}\\
\textbf{Method} & \textbf{R-1}  & \textbf{R-2} & \textbf{R-L} & \textbf{R-1}  & \textbf{R-2} & \textbf{R-L}\\
\midrule
SciBERT, w-1, l-1200 & 26.00 & 7.29 & 21.41 & 25.78 & 7.71 & 21.42 \\
SciBERT, w-1, l-1500 & 27.78 & 9.81 & 23.32 & 27.37 & 9.62 & 23.07 \\
SciBERT, w-1, l-1700 & 28.73 & 11.27 & 24.42 & 28.45 & 11.32 & 24.21\\
SciBERT, w-2, l-1700 & 30.15 & 12.32 & 25.66 & 29.71 & 12.04 & 25.35\\
SciBERT, w-5, l-1700 & 30.96 & 12.59 & 26.54 & 30.79 & 12.31 & 26.38\\
\bottomrule
\end{tabular}
\caption{Fine-tuning for PubHealth supervised multi-task model over positive sentence loss weight, base model and maximum length.}
\label{tab:finetune:ph:sup}
\end{table*}

\begin{table*}[t]
\centering
\fontsize{10}{10}\selectfont
\begin{tabular}{lrrrrrr}
\toprule
& \multicolumn{3}{c}{\bf Validation} & \multicolumn{3}{c}{\bf Test}\\
\textbf{Method} & \textbf{R-1}  & \textbf{R-2} & \textbf{R-L} & \textbf{R-1}  & \textbf{R-2} & \textbf{R-L}\\
\midrule
\multicolumn{7}{l}{\bf LIAR-PLUS Unsup}\\
Top-6  & 29.26 & 7.98 & 25.83 & 29.62 & 7.94 & 26.04\\ 
Filtered Top-6 & 29.52 & 7.90 & 25.98 & 29.60 & 7.96 & 25.94 \\
\midrule
\multicolumn{7}{l}{\bf LIAR-PLUS SUP}\\
Top-6  & 34.42 & 12.35 & 30.64 & 34.49 & 12.54 & 30.67 \\ 
Filtered Top-6 & 34.30 & 12.20 & 30.51 & 34.42 & 12.36 &  30.58 \\
\midrule
\multicolumn{7}{l}{\bf PubHealth Unsup}\\
Top-5 & 23.78 & 6.23 & 19.95 & 23.13 & 6.08 & 19.63 \\
Filtered Top-5 & 23.94 & 6.13 & 20.04 & 23.52 & 6.12 & 19.93 \\
\midrule
\multicolumn{7}{l}{\bf PubHealth SUP}\\
Top-5 & 30.24 & 12.61 & 26.36 & 29.78 & 12.50 & 26.18 \\
Filtered Top-5 & 30.35 & 12.63 & 26.43 & 29.93 & 12.42 & 26.24\\
\bottomrule
\end{tabular}
\caption{Sentence clean-up of long sentences for LIAR-PLUS and PubHealth.}
\label{tab:finetune:sentence-scores}
\end{table*}

\begin{table*}[t]
\centering
\fontsize{10}{10}\selectfont
\begin{tabular}{llll}
\toprule
\textbf{Method} & \textbf{Copy Rate} & \textbf{Novelty} & \textbf{Gold Coverage}\\ \midrule
\multicolumn{4}{l}{\bf LIAR-PLUS}\\
Top-6 Sup. & 100 & 0 & 29.2 $\pm$11.4\\
Justification & 41.4 $\pm$13.0 & 58.6 $\pm$13.0 & 100 \\ 
Top-6+Edits-6 Sup. & 98.5 $\pm$1.8 & 1.5 $\pm$1.8 & 30.7 $\pm$12.1\\
Top-6+Edits-6+Para Sup. &90.8 $\pm$4.8 & 9.2 $\pm$4.8 & 32.5 $\pm$12.6 \\
\midrule
\multicolumn{4}{l}{\bf PubHealth}\\
Top-5 Sup.& 100 & 0 & 26.3 $\pm$21.2 \\
Justification &  47.1 $\pm$21.0 & 52.9 $\pm$21.0 & 100 \\
Top-5+Edits-6 Sup.  & 98.1 $\pm$3.4 & 1.8 $\pm$2.0 & 27.8 $\pm$21.3\\
Top-5+Edits-6+Para Sup. & 90.4 $\pm$5.8  & 9.5 $\pm$5.2 & 28.5 $\pm$20.2\\
\bottomrule
\end{tabular}
\caption{Copy rate from the Ruling Comments, Novelty w.r.t the Ruling comments, and Coverage \% of words in the explanation that are found in the justification.}
\label{tab:results:novelty}
\end{table*}

\bgroup
\def\arraystretch{1.2}

\begin{table*}[t]
\centering
\fontsize{10}{10}\selectfont
\begin{tabular}{llrrrr}
\toprule
& \bf Method &  {\bf \small Flesch $\nearrow$} &  {\bf \small Flesch  CI} & {\bf \small Dale-Chall $\searrow$}  & {\bf \small Dale-Chall CI}  \\
\midrule
\multicolumn{6}{c}{\bf LIAR-PLUS}\\ 
\multirow{2}{*}{\bf Baselines} & Lead-4 & 50.89 & {\small [50.01-51.63]} & 8.75 & {\small  [8.71-8.80]} \\
& Lead-6 &  53.01 &  {\small  [52.41-53.64]} & 8.43  & {\small  [8.39-8.47]} \\
\hdashline
\multirow{3}{*}{\bf Supervised} & Top-6 (Supervised) & 57.77 &  {\small[57.15-58.38]} & 7.91 &  {\small[7.87-7.95]} \\
& Top-6+Para & 63.88 &  {\small[63.31-64.45]} & 7.55 & {\small [7.51-7.59]} \\
& Top-6+Edits-IE & 55.70 & {\small[55.03-56.36]} & 6.53 & {\small[6.50-6.56]}\\
& Top-6+Edits-IE+Edits-Gram & 59.52 &  {\small [58.89-60.17]} & 7.78  &   {\small[7.73-7.83]} \\
& Top-6+Edits+Para & 66.05 &  {\small [65.53-66.61]} & 7.46  & {\small [7.41-7.50]} \\
\hdashline
\multirow{3}{*}{\bf Unsupervised} & Top-6 (Unsupervised) & 52.84  & {\small [52.27-53.36]} & 8.52 &  {\small [8.48-8.55]}  \\
& Top-6+Para & 50.92 &  {\small [50.18-51.58]} &  6.97 &  {\small[6.94-7.01]} \\
& Top-6+Edits-IE & 50.70 & {\small [50.13-51.27]} & 6.92 &  {\small[6.89-6.94]} \\
& Top-6+Edits-IE+Edits-Gram & 54.76 &  {\small [54.15-55.34]} & 8.39 &  {\small [8.34-8.43]} \\
& Top-6+Edits-6+Para & 61.80 &  {\small [61.17-62.42]} & 8.01 &  {\small [7.97-8.05]} \\
\hdashline
 & \citet{atanasova-etal-2020-generating}-4 & 58.08 &  {\small[57.33-58.83]} & 8.03 & {\small [7.97-8.08]}\\
& Justification & 58.90 & {\small [58.23-59.68]} & 8.26 & {\small [8.20-8.32]}\\
\midrule
\multicolumn{6}{c}{\bf LIAR-PLUS Test Split Ablation} \\
\multirow{2}{*}{\bf Supervised} & Top-6+Para & 64.45 &  [63.81-65.04] & 7.52 &  [7.48-7.56] \\
& Top-6+Edits-IE & 56.26 & {\small [55.37-57.04]} & 6.51 &  {\small [6.48-6.55]} \\
\hdashline
\multirow{2}{*}{\bf Unsupervised} & Top-6+Para & 59.83 & {\small [59.20-60.36]} & 8.21 & {\small [8.16-8.25]} \\
& Top-6+Edits-IE & 59.34 & {\small [58.72-59.91]} & 8.14 & {\small [8.10-8.18]} \\
\midrule
\multicolumn{6}{c}{\bf PubHealth} \\
& Lead-3 & 44.76 & {\small [43.49-45.87]} & 9.12 & {\small [9.05-9.20]} \\
& Lead-5 & 46.00 & {\small [44.80-46.92]} & 8.88 & {\small [8.83-8.94]} \\
\hdashline
\multirow{3}{*}{\bf Supervised} & Top-5 (Supervised) & 49.56 &  {\small[48.73-50.27]} & 8.63 & {\small [8.58-8.68]}  \\
& Top-5+Para & 47.38 & {\small [46.50-48.15]} & 7.07 & {\small [7.04-7.12]}\\
& Top-5+Edits-IE & 57.53 & {\small [56.85-58.19]} & 8.18 & {\small [8.13-8.24]}\\
& Top-5+Edits-IE+Edits-Gram & 54.30 &  {\small [53.58-54.97]} & 8.33 & {\small [8.27-8.38]} \\
& Top-5+Edits-5+Para & 61.51 & {\small [60.89-62.19]} & 7.96 & {\small  [7.91-8.01]} \\
\hdashline
\multirow{3}{*}{\bf Unsupervised} & Top-5 (Unsupervised) & 43.55 &  {\small[42.51-44.52]} & 9.26 &  {\small[9.19-9.32] }\\
& Top-5+Para & 42.70 &  {\small[41.60-43.59]} & 7.35 & {\small [7.31-7.40]} \\
& Top-5+Edits-5-IE & 56.33 & {\small [55.68-56.97]} & 8.35 & {\small [8.31-8.40]} \\
& Top-5+Edits-5-IE+Edits-Gram & 50.46 & {\small [49.62-51.23]} & 8.65 &  {\small[8.59-8.70]} \\
& Top-5+Edits-5+Para & 60.25 & {\small [59.56-60.89]} & 8.13 &  {\small[8.08-8.19]} \\
\hdashline
& \citet{atanasova-etal-2020-generating}-3 & 49.69 & {\small [48.73-50.53]} & 8.81 & {\small [8.75-8.88]} \\
& Justification & 48.20 &  {\small[47.25-49.16]} & 9.22 &  {\small[9.15-9.32]} \\

\midrule
\multicolumn{6}{c}{\bf PubHealth - Test Split Ablation} \\
\multirow{2}{*}{\bf Supervised} & Top-5+Para & 46.23 & {\small [45.33-47.07]} & 7.11 & {\small [7.07-7.15]} \\
& Top-5+Edits-IE & 57.29 &  {\small[56.58-57.96]} & 8.21 &  {\small [8.16-8.26] } \\
\hdashline
\multirow{2}{*}{\bf Unsupervised} & Top-5+Para & 42.30 & {\small [41.34-43.28]} & 7.36 &  {\small [7.31-7.41]} \\
& Top-5+Edits-IE & 55.79 & {\small [55.16-56.44]} & 8.39 & {\small [8.34-8.43]} \\
\bottomrule
\end{tabular}
\caption{Readability measures (\S\ref{sec:result:readability}) over the validation splits. Readability measures include 95\% confidence intervals (\S\ref{sec:result:readability}, Metrics.). We report results reported from the prior work of \citet{atanasova-etal-2020-generating}-N, where we have the outputs to compute readability.}
\label{tab:results:readability:val}
\end{table*}

\begin{table*}[t]
\centering
\fontsize{10}{10}\selectfont
\begin{tabular}{llrrrrrrrr}
\toprule
& \bf Method & \textbf{R-1}$\nearrow$ & \textbf{R-1 CI}  & \textbf{R-2}$\nearrow$ & \textbf{R-2 CI} &  \textbf{R-L}$\nearrow$ & \textbf{R-L CI}  \\
\midrule
\multicolumn{9}{c}{\bf LIAR-PLUS}\\ 
\multirow{2}{*}{\bf Base.} & Lead-4 & 27.52 & {\small [26.99-28.00]} & 6.90 & {\small [6.54-7.30]} & 24.01 & {\small [23.53-24.49}] \\
& Lead-6 & 28.93 & {\small [28.39-29.43]} & 8.32 & {\small [7.92-8.76]} & 25.67 & {\small [25.15-26.16]} \\
\hdashline
\multirow{3}{*}{\bf Sup.} & Top-6 (Supervised) & 34.35 & {\small [33.71-34.94]} & 12.20 & {\small [11.72-12.70]} & 30.51 & {\small [29.97-31.09]}\\
& Top-6+Para & 34.51 & {\small[33.90-35.08]} & 11.49 & {\small[11.04-11.96]} & 30.68 & {\small[30.15-31.25]}\\
& Top-6+Edits-IE & 25.18 & {\small[24.63-25.72]} & 8.60 & {\small[8.23-8.98]} & 22.08 & {\small[21.58-22.55]}\\
& Top-6+Edits-IE+Edits-Gram & 34.07 & {\small [33.45-34.71]} & 11.58 & {\small [11.14-12.05]} & 30.12 & {\small [29.55-30.70]} \\
& Top-6+Edits-6+Para & 34.20 & {\small [33.62-34.78]} & 11.05 & {\small [10.59-11.46]} & 30.26 & {\small [29.71-30.83]}\\
\hdashline
\multirow{3}{*}{\bf Unsup.} & Top-6 (Unsupervised) & 29.29 & {\small [28.79-29.78]} & 7.99 & {\small [7.64-8.37]} & 25.84 & {\small [25.36-26.30]} \\
& Top-6+Para & 22.74 & {\small[22.31-23.24]} & 5.56 & {\small[5.29-5.82]} & 19.50 & {\small[19.06-19.93]}\\
& Top-6+Edits-IE & 21.46 & {\small[20.98-21.93]} & 5.67 & {\small[5.43-5.93]} & 18.76 & {\small[18.34-19.20]}\\
& Top-6+Edits-6+Edits-Gram & 29.02 & {\small [28.50-29.53]} & 7.47 & {\small [7.10-7.83]} & 25.52 & {\small [25.02-25.97]} \\
& Top-6+Edits-6+Para & 29.41 & {\small [28.89-29.90]} & 7.26 & {\small [6.90-7.60]} & 25.90 & {\small [25.43-26.37]}\\
\hdashline
 & \citet{atanasova-etal-2020-generating}-4 & 34.80 & {\small [34.13-35.39]} & 12.87 & {\small [12.29-13.46]} & 30.66 & {\small [30.08-31.26]} \\
\midrule

\multicolumn{9}{c}{\bf LIAR-PLUS Test Split Ablation}\\ 
\multirow{2}{*}{\bf Sup.} & Top-6+Para & 34.62 & {\small[34.02-35.23]} & 11.79 & {\small[11.33-12.21]} & 30.81 & {\small[30.26-31.35]}\\
& Top-6+Edits-IE & 25.48 & {\small[25.00-26.03]} & 8.75 & {\small[8.41-9.09]} & 22.29 & {\small[21.78-22.79]}\\
\hdashline
\multirow{2}{*}{\bf Unsup.} & Top-6+Para & 34.61 & {\small[34.08-35.19]} & 11.77 & {\small[11.30-12.26]} & 30.81 & {\small[30.20-31.38]}\\
& Top-6+Edits-IE & 25.48 & {\small[24.87-26.01]} & 8.75 & {\small[8.41-9.13]} & 22.29 & {\small[21.81-22.76]}\\
\midrule

\multicolumn{9}{c}{\bf PubHealth} \\
\multirow{2}{*}{\bf Base.}& Lead-3 & 23.18 & {\small [22.69-23.71]} & 5.55 & {\small [5.14-6.00]} & 18.74 & {\small [18.31-19.19]} \\
& Lead-5 & 23.31 & {\small [22.71-23.90]} & 6.07 & {\small [5.70-6.46]} & 19.60 & {\small [19.04-20.14]}\\
\hdashline
\multirow{3}{*}{\bf Sup.} & Top-5 (Supervised) & 30.37 & {\small [29.40-31.42]} & 12.64 & {\small [11.51-13.77]} & 26.46 & {\small [25.42-27.49]} \\
& Top-5+Para & 22.49 & {\small[21.62-23.39]} & 8.96 & {\small[8.22-9.75]} & 19.73 & {\small[18.82-20.64]}
\\
& Top-5+Edits-IE & 29.76 & {\small[28.81-30.70]} & 10.75 & {\small[9.91-11.63]} & 25.44 & {\small[24.56-26.32]}
\\
& Top-5+Edits-5+Edits-Gram & 29.62 & {\small [28.69-30.49]} & 11.20 & {\small [10.31-12.19]} & 25.54 & {\small [24.62-26.46]} \\
& Top-5+Edits-5+Para & 28.81 & {\small [27.97-29.70]} & 9.67 & {\small [8.94-10.37]} & 24.47 & {\small [23.71-25.31]}\\
\hdashline
\multirow{3}{*}{\bf Unsup.} & Top-5 (Unsupervised) & 23.80 & {\small [23.27-24.31]} & 5.76 & {\small [5.36-6.13]} & 19.24 & {\small [18.74-19.70]} \\
& Top-5+Para & 18.28 & {\small[17.64-18.88]} & 4.50 & {\small[4.22-4.79]} & 15.50 & {\small[14.90-16.14]}
\\
& Top-5+Edits-IE & 24.44 & {\small[23.85-25.04]} & 5.97 & {\small[5.67-6.31]} & 20.51 & {\small[19.98-21.01]}
\\
& Top-5+Edits-5+Edits-Gram & 23.74 & {\small [23.20-24.28]} & 5.72 & {\small [5.41-6.03]} & 19.76 & {\small [19.27-20.30]} \\
& Top-5+Edits-5+Para & 23.96 & {\small [23.49-24.49]} & 5.46 & {\small [5.17-5.78]} & 19.98 & {\small [19.52-20.45]} \\
\hdashline
& \citet{atanasova-etal-2020-generating}-3 & 31.05 & {\small [30.08-32.09]} & 12.66 & {\small [11.44-13.82]} & 26.45 & {\small [25.46-27.50]}\\

\midrule

\multicolumn{9}{c}{\bf PubHealth Test Split Ablation}\\ 
\multirow{2}{*}{\bf Sup.} & Top-5+Para & 22.09 & {\small[21.24-22.95]} & 8.75 & {\small[7.97-9.54]} & 19.48 & {\small[18.60-20.36]}
\\
& Top-5+Edits-IE & 29.46 & {\small[28.57-30.37]} & 10.71 & {\small[9.86-11.63]} & 25.53 & {\small[24.67-26.40]}
\\
\hdashline
\multirow{2}{*}{\bf Unsup.} & Top-5+Para & 18.11 & {\small[17.44-18.77]} & 4.41 & {\small[4.16-4.70]} & 15.48 & {\small[14.92-16.10]}
\\
& Top-5+Edits-IE & 18.11 & {\small[17.44-18.77]} & 4.41 & {\small[4.16-4.70]} & 15.48 & {\small[14.92-16.10]}
\\

\bottomrule
\end{tabular}
\caption{ROUGE-1/2/L F1 scores (\S\ref{sec:result:rouge}) of baselines (Base.), supervised (Sup.) and usupervised (Unsup.) methods over the validation splits. In \textit{italics}, we report results reported from prior work, where we do not always have the outputs to compute the confidence intervals.}
\label{tab:results:rouge:val}
\end{table*}

\bgroup
\def\arraystretch{1}

\end{document}